\documentclass[twoside]{article}

\usepackage{amssymb}
\usepackage{amsmath}

\usepackage{enumitem}
\usepackage{comment}
\usepackage{setspace}
\usepackage{textcase}

\usepackage{microtype}
\usepackage{graphicx}
\usepackage{wrapfig}
\usepackage{subcaption}
\usepackage{booktabs} % for professional tables

\usepackage{hyperref}

\usepackage{amsmath}
\usepackage{amssymb}
\usepackage{mathtools}
\usepackage{amsthm}

\usepackage{algorithm}
\usepackage[noend]{algpseudocode}

\usepackage[capitalize,noabbrev]{cleveref}

\usepackage{natbib}

\theoremstyle{plain}
\newtheorem{theorem}{Theorem}[section]
\newtheorem{proposition}[theorem]{Proposition}
\newtheorem{lemma}[theorem]{Lemma}

\theoremstyle{definition}
\newtheorem{definition}[theorem]{Definition}
\newtheorem{assumption}[theorem]{Assumption}
\theoremstyle{remark}

\newcommand{\BlackBox}{\rule{1.5ex}{1.5ex}}
\ifdefined\proof
  \renewenvironment{proof}{\par\noindent{\bf Proof\ }}{\hfill\BlackBox\\[2mm]}
\else
  \newenvironment{proof}{\par\noindent{\bf Proof\ }}{\hfill\BlackBox\\[2mm]}
\fi

\usepackage[textsize=tiny]{todonotes}
\usepackage[preprint]{aistats2026}
\begin{document}

% If your paper is accepted and the title of your paper is very long,
% the style will print as headings an error message. Use the following
% command to supply a shorter title of your paper so that it can be
% used as headings.
%
%\runningtitle{I use this title instead because the last one was very long}

% If your paper is accepted and the number of authors is large, the
% style will print as headings an error message. Use the following
% command to supply a shorter version of the author names so that
% they can be used as headings (for example, use only the surnames)
%
%\runningauthor{Surname 1, Surname 2, Surname 3, ...., Surname n}

\twocolumn[

  \aistatstitle{Theoretical Investigation on Inductive Bias of Isolation Forest}

  \begin{center}
    \textbf{Qin-Cheng Zheng}$ ^{1,2}$,~ \textbf{Shao-Qun Zhang}$ ^{1,3}$,~ \textbf{Shen-Huan Lyu}$ ^{4,5}$, \\
    \textbf{Yuan Jiang}$ ^{1,2}$,~ \textbf{Zhi-Hua Zhou}$ ^{1,2}$ \\
    $^1$ National Key Laboratory for Novel Software Technology, Nanjing University, China\\
    $^2$ School of Artificial Intelligence, Nanjing University, China \\
    $^3$ School of Intelligence Science and Technology, Nanjing University, China\\
    $^4$ College of Computer and Information, Hohai University, China\\
    $^5$ Key Laboratory of Water Resources Big Data Technology of Ministry of \\
    Water Resources, Hohai University, China
  \end{center}
  \aistatsaddress{}
]

\begin{abstract}
  Isolation Forest (iForest) stands out as a widely-used unsupervised anomaly detector, primarily owing to its remarkable runtime efficiency and superior performance in large-scale tasks. Despite its widespread adoption, a theoretical foundation explaining iForest's success remains unclear. This paper focuses on the inductive bias of iForest, which theoretically elucidates under what circumstances and to what extent iForest works well. The key is to formulate the growth process of iForest, where the split dimensions and split values are randomly selected. We model the growth process of iForest as a random walk, enabling us to derive the expected depth function, which is the outcome of iForest, using transition probabilities. The case studies reveal key inductive biases: iForest exhibits lower sensitivity to central anomalies while demonstrating greater parameter adaptability compared to $k$-Nearest Neighbor. Our study provides a theoretical understanding of the effectiveness of iForest and establishes a foundation for further theoretical exploration.
\end{abstract}

\section{INTRODUCTION} \label{section:introduction}

Unsupervised anomaly detection is a fundamental problem in the fields of machine learning and data mining. It is widely used in many real-world tasks, including fraud detection~\citep{Fawcett99Activity,Chandola09Anomaly}, network intrusion detection~\citep{Phoha02Internet}, and medical diagnosis~\citep{Wong03Bayesian,Fernando22Deep}. The goal is to identify anomalies in the concerned dataset, which are ``few and different'' from the normal samples~\citep{Liu08Isolation,Liu12Isolation}. Among various types of unsupervised anomaly detection methods, iForest~\citep{Liu12Isolation} stands out as one of the most popular choices. The basic idea is to gather a set of Isolation Trees (iTrees) trained from the observed data, and the anomaly score of a data point is defined as the average path length from the root node to the leaf node of the iTrees. The core concept of iTrees lies in uniformly randomly selecting an attribute and a split value, partitioning the data points into two subsets iteratively, and repeating this process until all data points are isolated. Although being highly heuristic, iForest is usually the first choice for anomaly detection. iForest outperforms other unsupervised anomaly detectors in many real-world applications in terms of accuracy, efficiency, and large-scale applicability~\citep{Liu08Isolation,Liu12Isolation,aggarwal2017introduction,pang2019deep,Cook20Anomaly,Pang21Deep}. 

Despite notable progress in practical applications, the theoretical understanding of the iForest algorithm remains limited. There are some seminal works that study topics such as convergence~\citep{Siddiqui16Finite}, PAC theory~\citep{Liu18Open}, and learnability~\citep{Fang22OOD} in the context of anomaly detection. However, these investigations are typically not specific to iForest and fail to provide a comprehensive explanation of the algorithm. \citet{Buschjager22Randomized} analyzed a modified version of iForest, whereas this variant deviates significantly from the original algorithm. To establish a comprehensive theoretical foundation for the success of iForest, it is crucial to focus directly on the algorithm itself and perform comparative analyses with other anomaly detection methods.

This paper aims to advance the theoretical understanding of iForest by investigating the conditions under which it works and the extent of its efficacy. A central focus of our study is to examine the inductive bias inherent in iForest. The algorithm's inherently stochastic growth mechanism, in which each split is selected randomly and can significantly alter the structure of an Isolation Tree, presents considerable challenges in evaluating its outcomes and identifying its inductive bias. To address this issue, we model the growth process of iTrees as a random walk, a special type of Markov chain. This perspective helps derive the closed-form expression of the expected depth function of iForest, the output of the iForest that serves as the criterion for anomaly detection. The expected depth function enables analysis of the inductive bias and provides a basic framework for understanding the success of iForest.

Our contributions can be summarized as follows:
\begin{enumerate}[topsep=0em,leftmargin=2em]
  \setlength{\itemsep}{0em}
  \item We are the first to open the black box of iForest by modeling the growth process as a random walk and derive the exact depth function, which supports the further analysis of inductive bias.
  \item We study the inductive bias of iForest by comparing it with \( k \)-NN through multiple case studies, concluding that iForest is less sensitive to central anomalies but more parameter-adaptive.
\end{enumerate}

The rest of this paper is organized as follows. Section~\ref{section:preliminary} introduces notations, settings, and concepts. Section~\ref{section:theory} presents the novel random walk view of the growth process of iForest and the derived exact depth function. Section~\ref{section:casestudy} analyzes the inductive bias of iForest through case studies. Section~\ref{section:empiricalstudies} conducts some empirical studies to verify the theoretical findings. Section~\ref{section:discussions} discusses the multi-dimensional scenarios. Section~\ref{section:conclusions} concludes this work. All proofs are deferred to the Appendix.

\section{Preliminary} \label{section:preliminary}

\noindent\textbf{Notations.}
We denote by \( f(n) = O(g(n)) \) and \( f(n) = \Omega(g(n)) \) if there exist constants \( c_1, c_2 > 0 \) such that \( f(n) \leq c_1 g(n) \) and \( f(n) \geq c_2 g(n) \), respectively, for all sufficiently large \( n \in \mathbb{N}^+ \). Similarly, \( f(n) = o(g(n)) \) and \( f(n) = \omega(g(n)) \) if \( f(n) / g(n) \to 0 \) and \( f(n) / g(n) \to \infty \) as \( n \to \infty \), respectively. Furthermore, \( f(n) = \Theta(g(n)) \) if both \( f(n) = O(g(n)) \) and \( f(n) = \Omega(g(n)) \) hold simultaneously. Let \( x_{i: j} \) denotes the vector \( (x_{i},\dots, x_{j}) \) for \( i \leq j \).

\textbf{Settings.}
Consider the unsupervised anomaly detection setting. Let $\mathcal{X} \subset \mathbb{R}^{d}$ and $\mathbf{x} = (\mathbf{x}^{(1)}, \dots, \mathbf{x}^{(d)}) \in \mathcal{X}$ be the input space and a $d$-dimensional vector, respectively. We observe $D = \{ \mathbf{x}_{1}, \mathbf{x}_{2}, \dots, \mathbf{x}_{n} \}$ consisting of $n$ samples drawn (allow non-i.i.d.) from an unknown distribution on \( \mathcal{X} \). Dataset $D$ contains $n_{0}$ normal samples and $n_{1}$ anomalies where $n = n_{0} + n_{1}$. There is an unknown and unlabeled anomaly subset $A \subsetneq D$, and we are required to find out the anomaly subset $A$.

In unsupervised anomaly detection tasks, however, ground-truth labels are unavailable, making the tasks inherently subjective. Different individuals may consider different anomaly subsets, depending on one's preferences. For such tasks, there are no loss functions to optimize, and understanding the inductive biases of various anomaly detection algorithms is crucial, rather than focusing solely on their absolute performance. In this study, we compare the inductive biases of two widely used anomaly detectors: iForest and $k$-NN.

% Like supervised learning, we have the ``No Free Lunch'' theorem for unsupervised anomaly detection.
% \begin{theorem}[No Free Lunch]
%   \label{theorem:nofreelunch}
%   Let $\mathcal{H}$ be a hypothesis space.
%   For any algorithm $\mathcal{A}$ and any dataset $D$, there exists a label sequence \( \{y_{i}\}_{i = 1}^{n} \) such that $\mathcal{L}(\mathcal{A}(D)) \geq 1/2$.
% \end{theorem}
% Theorem~\ref{theorem:nofreelunch} highlights a key insight: no single anomaly detection algorithm is guaranteed to perform well across all datasets, as there will always be some label assignment that leads to errors. 

\begin{algorithm}[t]
  \caption{BuildTree$(D)$} \label{algorithm:buildtree}
  \textbf{Input}: A dataset $D = \{ \mathbf{x}_{1}, \mathbf{x}_{2}, \dots \}$ \\
  \textbf{Output}: An Isolation Tree \( T \)
  \begin{algorithmic}[1]
    \If{$|D| \leq 1$}
    \State \textbf{return} Leaf
    \EndIf
    \State $j \gets$ uniform random in \(\{ j \mid \#\{\mathbf{x}^{(j)}\} > 1 \}\)
    \State $s \gets$ uniform random in \( [\min \mathbf{x}^{(j)}, \max \mathbf{x}^{(j)}] \)
    \State $D_{\text{left}} \gets \{ \mathbf{x} \in D \mid \mathbf{x}^{(j)} \leq s \}$
    \State $D_{\text{right}} \gets \{ \mathbf{x} \in D \mid \mathbf{x}^{(j)} > s \}$
    \State $\textrm{Node} \leftarrow \{ \text{SplitAtt} \leftarrow j$ \\
    $\quad \quad \quad \quad ~ \text{SplitValue} \leftarrow s$ \\
    $\quad \quad \quad \quad ~ \text{Left} \leftarrow \text{BuildTree}(D_{\text{left}})$ \\
    $\quad \quad \quad \quad ~ \text{Right} \leftarrow \text{BuildTree}(D_{\text{right}}) \}$
    \State \textbf{return} $\textrm{Node}$
  \end{algorithmic}
\end{algorithm}
\begin{algorithm}[t]
  \caption{Depth $(\mathbf{x}, T)$} \label{algorithm:pathlength}
  \textbf{Input}: A sample $\mathbf{x}$ and an Isolation Tree $T$ \\
  \textbf{Output}: The depth of $\mathbf{x}$ in $T$
  \begin{algorithmic}[1]
    \If {$T$ is a Leaf}
    \State \textbf{return} $0$
    \EndIf
    \State $j \gets T.\text{SplitAtt}, s \gets T.\text{SplitValue}$
    \If {$\mathbf{x}^{(j)} \leq s$}
    \State \textbf{return} $1 + \text{Depth}(\mathbf{x}, T.\text{Left})$
    \Else
    \State \textbf{return} $1 + \text{Depth}(\mathbf{x}, T.\text{Right})$
    \EndIf
  \end{algorithmic}
\end{algorithm}
\begin{algorithm}[t]
  \caption{Isolation Forest} \label{algorithm:isolationforest}
  \textbf{Input}: A dataset $D$ and the tree number $M$ \\
  \textbf{Output}: A score function
  \begin{algorithmic}[1]
    \For{$m = 1, \dots, M$}
    \State $h_{m} \gets \textrm{Depth}(\cdot, \textrm{BuildTree}(D))$
    \EndFor
    \State \textbf{return} $M^{-1} \sum_{m = 1}^M h_{m}$
  \end{algorithmic}
\end{algorithm}

\textbf{iForest.} The concept of iForest~\citep{Liu08Isolation} has exceeded the scope of anomaly detection and become a learning framework for diverse machine learning tasks, including density estimation~\citep{Ting21Density}, time-series analysis~\citep{Ting21Isolation}, and so on. There are also many variants of iForest, such as SCiForest~\citep{Liu10SCiForest}, LSHiForest~\citep{Zhang17LSHiForest}, EIF~\citep{Hariri21EIF}, Deep Isolation Forest~\citep{Xu23DeepiForest}, etc. Note that theoretically understanding the large family of isolation-based methods is a long way to go. This paper focuses on the original iForest~\citep{Liu08Isolation}, which is the basis of understanding the idea of isolation as well as the large algorithm family.

iForest works under the belief that anomalies are easier to isolate than normal data points if we uniformly randomly partition the whole feature space until all the points are isolated. iForest constructs a collection of iTrees independently, and the anomaly score is negatively correlated with the average depth, the path length from the root node to the leaf node. The procedure of iForest is detailed in Algorithms~\ref{algorithm:buildtree}-\ref{algorithm:isolationforest}.

For convenience, we introduce two simplifications: relaxing the maximum depth of iTrees and directly outputting the average depth~\citep{Liu08Isolation} instead of computing the anomaly score. About the former simplification, constraining the maximum depth will not alter the relative order of isolation depths for different data points and is usually applied to detect multiple anomaly candidates~\citep{Liu08Isolation}, which is not the focus of this work. As for the latter simplification, note that the anomaly score introduced in~\citet{Liu08Isolation}
\[
  s(\mathbf{x}, n) = 2^{-\mathbb{E}_{\mathbf{\Theta}} [h(\mathbf{x}; D, \mathbf{\Theta})] / c(n)}
\]
constitutes a monotonically decreasing transformation of the average depth. This shifts the focus from analyzing maximum anomaly scores to minimum depths without altering the concerned problem.

\textbf{$k$-NN anomaly detectors.} Nearest neighbor-based methods can be roughly separated into two categories: density-based methods that assume the anomalies are in low-density regions and distance-based methods that assume the anomalies are far from normal points. Among the former, the most popular ones include Local Outlier Factor~\citep{Breunig00LOF}, and its variants~\citep{Tang02LOF,Papadimitriou03LOCI,Fan06ROF}. Among the latter, the most typical ones define the anomaly score as the distance to the $k$-th nearest neighbors of a point~\citep{Byers98NearestNeighbor,Guttormsson99Elliptical}.
% CAN LEAVE OR REMOVE: 
Here, we consider the most standard $k$-NN, which defines the anomaly score as the average distance to the $k$ nearest neighbors. Formally, the scoring function is defined by
\begin{equation*}
  h_{knn}(\mathbf{x}; D) \triangleq \frac{1}{k} \sum_{\mathbf{x}^{\prime} \in \mathcal{N}_{k}(\mathbf{x})} \| \mathbf{x} - \mathbf{x}^{\prime} \|_{1} \ ,
\end{equation*}
where $\mathcal{N}_{k}(\mathbf{x})$ is the set of the $k$ nearest neighbors of $\mathbf{x}$ in $D$. The choice of the $L_{1}$-norm is representative because all norms in finite-dimensional spaces are equivalent. The reason we select the $L_{1}$-norm is that each iTree partitions the entire feature space into multiple hyper-rectangles. This characteristic aligns more closely with the $L_{1}$-norm than the more commonly used $L_{2}$-norm.

Here, we employ the $k$-NN algorithm rather than density-based approaches. It becomes apparent that density-based algorithms employing the rectangle kernel \( K(u) = 2^{-d} ~ \mathbb{I}(\| u \|_{1} \leq 1) \) tend to be equivalent to $k$-NN. This equivalence arises due to the strong correlation between the number of points in the neighbor of one point and the distance to its neighbors.
\section{DEPTH FUNCTION OF IFOREST} \label{section:theory}

In this section, we mathematically analyze the depth function, which is the scoring criterion output by iForest, and serves as a basis for inductive bias analysis. We model the growth of iTree as a random walk and derive the closed-form expression of depth functions in the one-dimensional case. % Given the random walk model, we derive the closed-form expression of the expected depth function, which represents the first theoretical result for the outcome of iForest.

\subsection{The random walk model for iTrees} \label{subsection:random_walk_model}

We denote the average depth of a point $\mathbf{x}$ in an iTree as $h(\mathbf{x}; D, \mathbf{\Theta})$, where $\mathbf{\Theta}$ consists of all the randomness from the construction of the iTree, including the randomness of attribute selections and split values sampling. To begin with, we introduce the following conclusion.
\begin{proposition}[Concentration of iForest] \label{proposition:concentration}
  For any fixed dataset $D$ and any $\mathbf{x} \in \mathcal{X}$, we have
  \begin{align*}
    \Pr \Bigg[
    \Bigg|
    \frac{1}{M} \sum_{m = 1}^{M} h(\mathbf{x}; D, \mathbf{\Theta}_{m}) - & \mathbb{E}_{\mathbf{\Theta}} [ h(\mathbf{x}; D, \mathbf{\Theta})] \Bigg| \geq \epsilon \Bigg] \\
                                                                         & \  \leq 2 \exp \left( - 2 \epsilon^{2} M / n^{2} \right) \ ,
  \end{align*}
  where $h(\mathbf{x}; D, \mathbf{\Theta}_{m})$ is the depth of $\mathbf{x}$ in the $m$-th tree.
\end{proposition}
\noindent Proposition \ref{proposition:concentration} shows that the empirical mean of the depth function $M^{-1} \sum_{m = 1}^{M} h(\mathbf{x}; D, \mathbf{\Theta}_{m})$ will with high probability converges to its expectation $\mathbb{E}_{\mathbf{\Theta}} [h(\mathbf{x}; D, \mathbf{\Theta})]$ as the number of trees $M$ increases. Following the proposition, it suffices to analyze the expected depth function. However, the randomness of the generating process of iTrees poses a challenge, as every split is chosen randomly but depends on the dataset, and a different split can lead to a completely different tree.

Next, we demonstrate the random walk model of the growth process of iTrees. We begin with the case of $d = 1$, i.e., $x_{i} \in \mathcal{X} = \mathbb{R}$. Without loss of generality, we assume that $x_{1} < x_{2} < \dots < x_{n}$ is sorted in ascending order. Suppose we care about the depth of $x_{i}$, each tree node forms an interval, and let $x_{\ell_{t}}$ and $x_{r_{t}}$ denote the endpoints of the interval containing $x_{i}$ at time $t$, respectively. The following theorem shows that $\mathbf{s}_{t} = (x_{\ell_{t}}, x_{r_{t}})$ by the growth of iTrees is a random walk.

\begin{theorem}[Random Walk Model for iTree] \label{theorem:random_walk_model}
  For any $x_{\ell}, x_{r}, x_{\ell^{\prime}}, x_{r^{\prime}} \in D$, random process $\mathbf{s}_{t}(x_{i}) \triangleq(x_{\ell_{t}}, x_{r_{t}})$ is a random walk with transition probability
  \begin{align}
     & \Pr[\mathbf{s}_{t + 1} = (x_{\ell^{\prime}}, x_{r^{\prime}}) \mid \mathbf{s}_{t} = (x_{\ell}, x_{r})] \nonumber                                                                                                                          \\
     & \quad\quad = \left\{ \begin{aligned}
                                    & 1,                                                                     & \text{if } \ell^{\prime} = \ell = i = r^{\prime} = r \ , \\
                                    & \frac{x_{\ell^{\prime}} - x_{\ell^{\prime} - 1}}{x_{r} - x_{\ell}} \ , & \text{if } \ell^{\prime} > \ell, r^{\prime} = r \ ,      \\
                                    & \frac{x_{r^{\prime} + 1} - x_{r^{\prime}}}{x_{r} - x_{\ell}} \ ,       & \text{if } \ell^{\prime} = \ell, r^{\prime} < r \ ,      \\
                                    & 0 \ ,                                                                  & \text{otherwise} \ .
                                 \end{aligned} \right. \label{equation:random_walk_transition_probability}
  \end{align}
\end{theorem}

\begin{figure}[t]
  \centering
  \begin{minipage}[htbp]{0.98\columnwidth}
    \centering
    \includegraphics[trim=0 4 0 0,clip,width=0.95\linewidth]{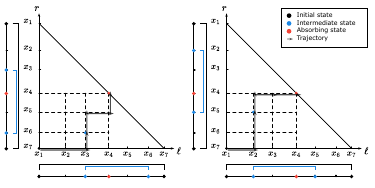}
    \caption{Examples of the random walk model for iTrees.
      The points colored in black, blue, and red indicate the initial, intermediate, and absorbing states, respectively. The lines with arrows indicate the trajectory of the transition process.}
    \label{fig:random_walk}
  \end{minipage}
  \hfill
\end{figure}

Figure~\ref{fig:random_walk} shows the random walk model for iTrees given a seven point dataset $D = \{ x_{1}, x_{2}, \dots, x_{7} \}$, of which the relative positions are shown in the two axes. The left and right subfigures are two possible trajectories of the random walk. Every state of the random walk can be represented by a point in the two-dimensional coordinate system, which represents the leftmost and rightmost points in the tree node containing the concerned point. The initial state of the random walk is $\mathbf{s}_{0} = (x_{1}, x_{7})$, which is colored in black. At each step, it is allowed to move to the rightward or upward, keeping to the left and below the absorbing state $(x_{4}, x_{4})$, which is colored in red. Moving rightward and upward mean split at the left of $x_{i}$ and the right of $x_{i}$, respectively. The probability of moving how large the step size is proportional to the distance between two consecutive points, as defined in Eq.~\eqref{equation:random_walk_transition_probability}. For example, as shown in the left of Figure~\ref{fig:random_walk}, the random walk reaches the state $(x_{3}, x_{6})$ starting from $(x_{1}, x_{7})$ by moving rightward (with probability $\frac{x_{3} - x_{2}}{x_{7} - x_{1}}$) and then upward (with probability $\frac{x_{7} - x_{6}}{x_{7} - x_{3}}$). The random walk keeps running until it reaches the absorbing state.

Though we only show the random walk model when $d = 1$, the model still holds for any $d > 1$. This is because the candidate split attribute and split value only depend on the current candidate points, no matter how the current state is reached, i.e., the growth of iTrees verifies the Markov property.

\subsection{Expected depth of a data point} \label{subsection:expected_depth}

With the random walk model at hand, we are now possible to analyze the expected depth function. We begin with a simpler notation as follows
\begin{equation*}
  \bar{h}(x_{i}; x_{1}, \dots, x_{n}) \triangleq \mathbb{E}_{\mathbf{\Theta}} [h(x_{i}; x_{1}, \dots, x_{n}, \mathbf{\Theta})] \ ,
\end{equation*}
where $x_{i} \in D$. We first introduce the following lemma, which can simplify the analysis.

\begin{lemma}
  \label{lemma:depth_decomposition}
  For any given dataset $D$ with sample size $n \geq 3$ and $x_{i} \in D$ for $1 < i < n$, we have
  \begin{multline*}
    \bar{h} \left( x_{i}; x_{1}, \dots, x_{n} \right) \\
    = \bar{h} \left( x_{i}; x_{1}, \dots, x_{i} \right) + \bar{h} \left( x_{i}; x_{i}, \dots, x_{n} \right) \ .
  \end{multline*}
\end{lemma}

\noindent Lemma~\ref{lemma:depth_decomposition} shows that the expected depth of a data point $x_{i}$ can be decomposed into two parts, of which the concerned data point is the rightmost and the leftmost, respectively.
From the random walk perspective, the number of steps required to reach the absorbing point $(x_{i}, x_{i})$ equals the sum of the steps taken in the rightward direction and the steps taken upward independently. To be more specific, although opting to move rightward can increase the probability of moving upward in the next step, the probability distribution of the step size of moving, conditioning on moving upward before, remains unchanged; thus, the expected total steps remain unchanged, too.

% We first write the transition matrix of the random walk model in Eq.~\ref{equation:random_walk_transition_probability} as
% \begin{equation*}
%   \mathbf{P} = \begin{bmatrix}
%    0      & \frac{x_{(2)} - x_{(1)}}{x_{(i)} - x_{(1)}} & \frac{x_{(3)} - x_{(2)}}{x_{(i)} - x_{(1)}} & \dots  & \frac{x_{(i)} - x_{(i - 1)}}{x_{(i)} - x_{(1)}} \\
%     0      & 0                                           & \frac{x_{(3)} - x_{(2)}}{x_{(i)} - x_{(2)}} & \dots  & \frac{x_{(i)} - x_{(i - 1)}}{x_{(i)} - x_{(2)}} \\
%     0      & 0                                           & 0                                           & \dots  & \frac{x_{(i)} - x_{(i - 1)}}{x_{(i)} - x_{(3)}} \\
%     \vdots & \vdots                                      & \vdots                                      & \ddots & \vdots                                          \\
%     0      & 0                                           & 0                                           & \dots  & 1
%   \end{bmatrix} \ .
% \end{equation*}

From the above analysis, it suffices to analyze the expected depth of the rightmost point $x_{i}$ in $\{ x_{1}, \dots, x_{i} \}$, as the leftmost point can be analyzed similarly. By the transition rule of random walks, the depth of point $x_{i}$ has a cumulative distribution function as follows.
\begin{multline*}
  \Pr_{\mathbf{\Theta}}[
    h(x_{i}; x_{1}, \dots, x_{i}, \mathbf{\Theta}) \leq \xi
  ] \\
  = [1, 0, \dots, 0] \cdot \mathbf{P}^{\xi} \cdot [0, 0, \dots, 1]^{\top} \ ,
\end{multline*}
where $\mathbf{P} = [p_{jk}]_{i \times i}$ is the transition matrix with
\begin{equation*}
  p_{jk} = \Pr[\mathbf{s}_{t + 1} = (x_{k}, x_{i}) \mid \mathbf{s}_{t} = (x_{j}, x_{i})]
\end{equation*}
and $\xi \in \mathbb{N}^{+}$.
Thus, the expected depth of $x_{i}$ equals
\begin{equation*}
  \bar{h}(x_{i}) = \sum_{\xi = 1}^{i} \Pr_{\mathbf{\Theta}}[h(x_{i}; \mathbf{\Theta}) \leq \xi] \ ,
\end{equation*}
where $\bar{h}(x_{i}; x_{1}, \dots, x_{i})$ and \( h(x_{i}; x_{1}, \dots, x_{i}) \) are abbreviated as $\bar{h}(x_{i})$ and \( h(x_{i}) \), respectively. By some algebraic manipulations, we have the closed-form expression of the expected depth function below.
\begin{lemma}
  \label{lemma:expected_depth}
  For any given dataset $D = \{ x_{1}, \dots, x_{i} \}$ with sample size $i > 1$, we have
  \begin{align*}
    \bar{h}(x_{i}; x_{1}, \dots, x_{i}) = \sum_{j = 2}^{i} \frac{x_{j} - x_{j - 1}}{x_{i} - x_{j - 1}} \ , \\
    \bar{h}(x_{1}; x_{1}, \dots, x_{i}) = \sum_{j = 2}^{i} \frac{x_{j} - x_{j - 1}}{x_{j} - x_{1}} \ .
  \end{align*}
\end{lemma}
\noindent Lemma~\ref{lemma:expected_depth} presents the closed-form expression of the expected depths of the rightmost or leftmost point. Based on a straightforward combination of Lemma~\ref{lemma:depth_decomposition} and Lemma~\ref{lemma:expected_depth}, we have the expected depth of any $x_{i}$ in $D$ in the following theorem.

\begin{theorem}
  \label{theorem:expected_depth}
  For any given dataset $D$ with sample size $n > 2$ and $x_{i} \in D$, we have
  \begin{equation*}
    \bar{h}(x_{i}; x_{1}, \dots, x_{n}) = \sum_{j = 2}^{i} \frac{x_{j} - x_{j - 1}}{x_{i} - x_{j - 1}} + \sum_{j = i + 1}^{n} \frac{x_{j} - x_{j - 1}}{x_{j} - x_{i}} \ .
  \end{equation*}
\end{theorem}
\noindent Theorem~\ref{theorem:expected_depth} presents the closed-form expression for the depth function of any $x_{i} \in D$. This formula represents the first theoretical characterization of the decision function output by iForest, providing a foundational basis for further analysis of the inductive bias.

For any unseen test data points, the depth function is the linear interpolation of the depths of the two nearest data points. This is a byproduct of this paper and is beyond the scope of inductive bias analysis. For anyone interested, we refer the reader to Theorem~\ref{theorem:expected_depth_outer}, where we provide the detailed conclusion and proof.

\section{INDUCTIVE BIAS} \label{section:casestudy}

Based on the closed-form expression of expected depth functions, we now conduct case studies to compare the inductive bias of iForest with $k$-NN. Before that, we define the density metrics.

\begin{definition}
  \label{definition:density_metrics}
  Let $U \triangleq \max_{i \geq 2} ~ \lvert x_{i + 1} - x_{i} \rvert$ and $L \triangleq \min_{i \geq 2} ~ \lvert x_{i + 1} - x_{i} \rvert$ be the maximum and minimum distances between adjacent points, respectively. Define two density metrics as follows:
  \begin{itemize}[topsep=0em,leftmargin=2em]
    \item Density ratio: $\kappa = U / L$.
    \item Density difference: $\delta = U - L$.
  \end{itemize}
\end{definition}
The two metrics evaluate how non-uniformly data are distributed relative to their neighbors, measured by ratio and difference. Intuitively, a dataset composed entirely of normal points is expected to exhibit small values for both metrics; otherwise, some points would be excessively distant from their neighbors. In the case studies below, we show that these density metrics play a critical role in shaping the inductive bias of iForest.

Before proceeding, we introduce the assumption.
\begin{assumption}
  \label{assumption:density_ratio}
  Assume that \( \kappa \geq \sqrt{n + 3} \).
\end{assumption}
\noindent Assumption~\ref{assumption:density_ratio} is frequently observed in practice. To justify its reasonability, we check all dimensions of binary classification datasets from the OpenML benchmark~\citep{Vanschoren14OpenML,Bischl21OpenML,Bischl25OpenML}. The results are summarized in Table~\ref{table:density_ratio_verification}. ``Successful'' indicates that the density ratio exceeds $\sqrt{n + 3}$. ``Valid'' indicates the nonexistence of repeated values in the dimension. Notably, the theoretical probability of repeated values is zero. Thus, we omit this case for the convenience of analysis. Our analysis demonstrates that the density ratio exceeds $\sqrt{n + 3}$ in almost all dimensions across all datasets, highlighting the mildness of Assumption~\ref{assumption:density_ratio}. A theoretical study of the mildness is also conducted, with detailed results in Appendix~\ref{section:elaboration_assumption_kappa}.

\begin{table}[ht]
  \caption{Number of features in OpenML benchmark that successfully verify Assumption~\ref{assumption:density_ratio}.}
  \label{table:density_ratio_verification}
  \vspace{4pt}
  \centering
  \begin{tabular}{ccc}
    \toprule
    \textbf{Successful} & \textbf{Valid} & \textbf{Total} \\
    \midrule
    930,738             & 930,751        & 933,440        \\
    \bottomrule
  \end{tabular}
\end{table}

\subsection{Marginal single anomaly}
\label{subsection:casestudy_marginal_single_anomalies}

\begin{figure}[ht]
  \centering
  \includegraphics[width=0.45\textwidth]{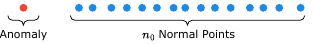}
  \caption{Marginal single anomaly.}
  \label{fig:casestudy_marginal_single_anomaly}
\end{figure}

The first case study is about the marginal single anomaly, an example of which is shown in Figure~\ref{fig:casestudy_marginal_single_anomaly}, where the anomaly is located at the leftmost, and the normal points are distributed on the right side. For the right-most anomaly with normal points on the left, we can have a similar conclusion. Marginal single anomalies may be the most common ones in practice, which happens when there is some abrupt change in normal data drawn from continuous and bounded support. Note that although this model looks simple, it is the basic building block of more complex models.

For the marginal single anomaly, we have the following theorem for iForest.
\begin{theorem}
  \label{theorem:marginal_single_anomaly}
  Suppose that $x_{2} - x_{1} > U \cdot \kappa$. For any dataset $D$ with $x_{2:n}$ verifying $\kappa$-dense, it holds that $\bar{h}(x_{1}; x_{1: n}) < \bar{h}(x_{j}; x_{1: n})$ for all $j > 1$.
\end{theorem}
\noindent Theorem~\ref{theorem:marginal_single_anomaly} shows that the expected depth of the marginal single anomaly is the smallest among all the data points when \( x_{2} - x_{1} > U \cdot \kappa \).
Hence, the marginal single anomaly is the most likely to be the anomaly output by iForest.
For the rightmost marginal single anomaly, Theorem~\ref{theorem:marginal_single_anomaly} holds similarly. Theorem~\ref{theorem:marginal_single_anomaly} shows the sufficiency of $x_{2} - x_{1} > U \cdot \kappa$. The following theorem shows the necessity.

\begin{theorem}
  \label{theorem:marginal_single_anomaly_necessity}
  Suppose that $U < x_{2} - x_{1} \leq U \cdot \kappa$, and $x_{2:n}$ verifies $\kappa$-dense.
  For any $n > 4$, there exists an assignment to $x_{1}, \dots, x_{n}$ such that $\bar{h}(x_{1}; x_{1: n}) \geq \bar{h}(x_{j_{0}}; x_{1: n})$ holds for some $j_{0} > 1$.
\end{theorem}
\noindent Theorem~\ref{theorem:marginal_single_anomaly_necessity} shows that if $x_{2} - x_{1} > U \cdot \kappa$ is violated, even though $x_{2} - x_{1}$ is the largest distance between two consecutive points, it remains possible for $x_{1}$ not to be the shallowest point, or in other words, iForest may fail to detect the marginal single anomaly.

For $k$-NN, we have the following conclusion.
\begin{theorem} \label{theorem:marginal_single_anomaly_nearest_neighbor}
  Suppose that $x_{2: n}$ verifies $\delta$-dense.
  The sufficient and necessary condition for $k$-NN to detect the marginal single anomaly is $x_{2} - x_{1} > U + (k - 1)\delta / 2$.
\end{theorem}
\noindent Theorem~\ref{theorem:marginal_single_anomaly_nearest_neighbor} reveals that the threshold for $k$-NN depends on the choice of $k$; when $k$ is too large, the marginal single anomaly is likely to be missed, while choosing a small $k$ may ignore some global information.
In summary, Theorems~\ref{theorem:marginal_single_anomaly}-\ref{theorem:marginal_single_anomaly_nearest_neighbor} imply that isolation-based anomaly detection is more adaptive to marginal single anomalies than $k$-NN.

\subsection{Central single anomaly}
\label{subsection:casestudy_central_single_anomaly}

\begin{figure}[ht]
  \centering
  \includegraphics[width=0.45\textwidth]{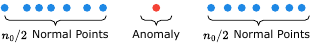}
  \caption{Central single anomaly.}
  \label{fig:casestudy_central_single_anomaly}
\end{figure}

Central single anomaly is another common type of anomaly, as shown in Figure~\ref{fig:casestudy_central_single_anomaly}. The anomaly is located at the center of the data points, and the normal points are distributed on both the left and right sides. This models the scenario when the data distribution consists of multiple clusters, and the anomaly is likely to arise in the gap between different clusters.

For iForest, we have the following theorem.
\begin{theorem}
  \label{theorem:central_single_anomaly}
  Let \( n_{0} \) be an even number for convenience.
  Suppose that both $x_{1: n_{0} / 2}$ and $x_{(n_{0} / 2 + 2): n}$ verify $\kappa$-dense.
  Then, the sufficient and necessary condition for iForest to detect $x_{n_{0} / 2 + 1}$ as an anomaly is that $\min \{ x_{n_{0} / 2 + 1} - x_{n_{0} / 2}, x_{n_{0} / 2 + 2} - x_{n_{0} / 2 + 1} \} > \Theta(\sqrt{n_{0}\kappa})$.
\end{theorem}
\noindent Theorem~\ref{theorem:central_single_anomaly} shows the sufficiency and necessity for iForest detecting the central single anomaly. Here, the anomaly is ``different'' enough from the normal points to a degree of order $\Theta(\sqrt{n_{0}\kappa})$, which is a relatively large quantity, especially when $n_{0}$ is large. This may be because iForest is more likely to capture global information for data distribution.

For $k$-NN, we have the following theorem.
\begin{theorem}
  \label{theorem:central_single_anomaly_nearest_neighbor}
  Let $n_{0}$ be an even number for convenience. Suppose that both $x_{1: n_{0} / 2}$ and $x_{(n_{0} / 2 + 2): n}$ verify $\delta$-dense. The sufficient and necessary condition for $k$-NN to detect the central single anomaly is that $\min \{ x_{n_{0} / 2 + 1} - x_{n_{0} / 2}, x_{n_{0} / 2 + 2} - x_{n_{0} / 2 + 1} \} > \Theta(k \delta)$.

\end{theorem}
\noindent Theorem~\ref{theorem:central_single_anomaly_nearest_neighbor} shows that the threshold for $k$-NN also relies on the selection of $k$, which is an algorithm-dependent parameter, while the threshold for iForest relies only on the data distributions, which is problem-dependent. We conclude that iForest is more careful about detecting central single anomalies but is also more parameter-adaptive than $k$-NN.

\subsection{Marginal clustered anomalies}
\label{subsection:casestudy_marginal_clustered_anomalies}

\begin{figure}[ht]
  \centering
  \includegraphics[width=0.45\textwidth]{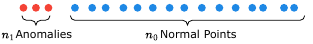}
  \caption{Marginal multiple anomalies.}
  \label{fig:casestudy_marginal_multiple_anomalies}
\end{figure}

In this subsection, we consider the case of marginal clustered anomalies. The motivation is that anomaly events may occur multiple times, as shown in Figure~\ref{fig:casestudy_marginal_multiple_anomalies}. A group of anomalies is located at the leftmost, and the normal points are distributed on the right. Marginal clustered anomalies characterize some unimodal data with multiple clustered noisy points.

For marginal clustered anomalies, we have the following conclusion for iForest.
\begin{theorem}
  \label{theorem:marginal_multiple_anomalies}
  Let \( n_{1} \) be an odd number for convenience where \(n_{1} = o(n_{0}) \). For any dataset \(D\), of which \( x_{1: n_{1}} \) and \( x_{(n_{1} + 1): (n_{1} + n_{0})} \) verify \( \kappa \)-dense, the sufficient and necessary condition for iForest to detect all the marginal clustered anomalies is \( x_{n_{1} + 1} - x_{n_{1}} > \Theta(n_{1}^{2} \kappa) \).
\end{theorem}
\noindent Theorem~\ref{theorem:marginal_multiple_anomalies} shows that iForest can detect all the clustered anomalies when normal points are far enough from anomalies, while it requires a quite large threshold of order \(\Theta(n_{1}^{2} \kappa)\) when \( n_1 \) is large.

For $k$-NN, we have the following theorem.
\begin{theorem}
  \label{theorem:marginal_multiple_anomalies_nearest_neighbor}
  Suppose \( \omega(n_{1}) \leq k \leq o(n_{0})  \). $x_{1:n_{1}}$ and $x_{(n_{1} + 1): (n_{1} + n_{0})}$ verify $\delta$-dense. The sufficient and necessary condition for $k$-NN to detect all marginal clustered anomalies is $x_{n_{1} + 1} - x_{n_{1}} > \Theta(k \delta)$.
\end{theorem}
\noindent Theorem~\ref{theorem:marginal_multiple_anomalies_nearest_neighbor} shows that the threshold for $k$-NN is also algorithm-dependent, yielding properties similar to those in Theorems~\ref{theorem:marginal_single_anomaly_nearest_neighbor} and~\ref{theorem:central_single_anomaly_nearest_neighbor}.
The selection of $k$ holds considerable importance for the $k$-NN.
For instance, when $k$ is too small, $k$-NN may wrongly label the clustered anomalies as normal points, leading to undesirable missed detections and false positives.
In contrast, iForest keeps exhibiting adaptability without any hyperparameter tuning. Note that the parameters of iForest are often set to trade off between the approximation performance and the computational cost, while the parameters of $k$-NN are set to trade off between the false positives and the false negatives.

\begin{table}[htbp]
  \centering
  \caption{Decision thresholds of iForest and $k$-NN for different types of anomalies.}
  \vspace{4pt}
  \label{table:thresholds}
  \begin{tabular}{lcc}
    \toprule
    \textbf{Anomaly type} & \textbf{iForest}              & \textbf{$k$-NN}    \\
    \midrule
    Marginal single       & $U \cdot \kappa$              & $\Theta(k \delta)$ \\
    Central single        & $\Theta(\sqrt{n_{0} \kappa})$ & $\Theta(k \delta)$ \\
    Marginal clustered    & $\Theta(n_{1}^{2} \kappa)$    & $\Theta(k \delta)$ \\
    \bottomrule
  \end{tabular}
\end{table}

Table~\ref{table:thresholds} summarizes three cases of anomalies in this section. It is observed that iForest acts to be more cautious in reporting an anomaly as the threshold is relatively large, but more parameter-adaptive to different types of anomalies than $k$-NN, as the thresholds of iForest are only problem-dependent.
\begin{figure}[ht]
  \centering
  \begin{subfigure}[b]{0.48\columnwidth}
    \centering
    \includegraphics[width=\textwidth]{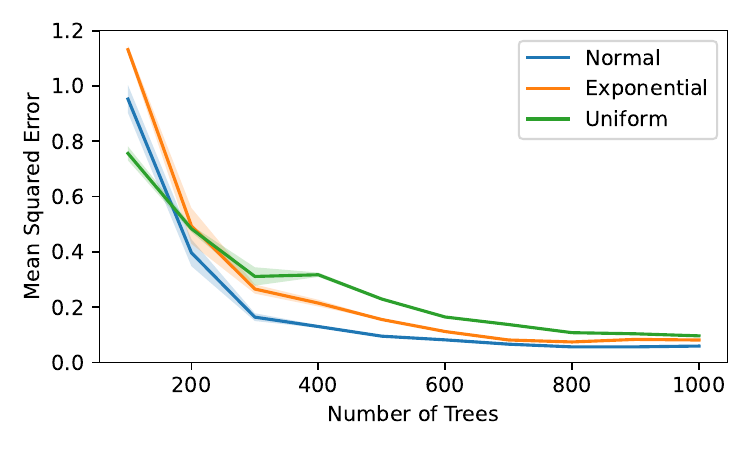}
    \subcaption{Synthetic datasets}
  \end{subfigure}
  \hfill
  \begin{subfigure}[b]{0.48\columnwidth}
    \centering
    \includegraphics[width=\textwidth]{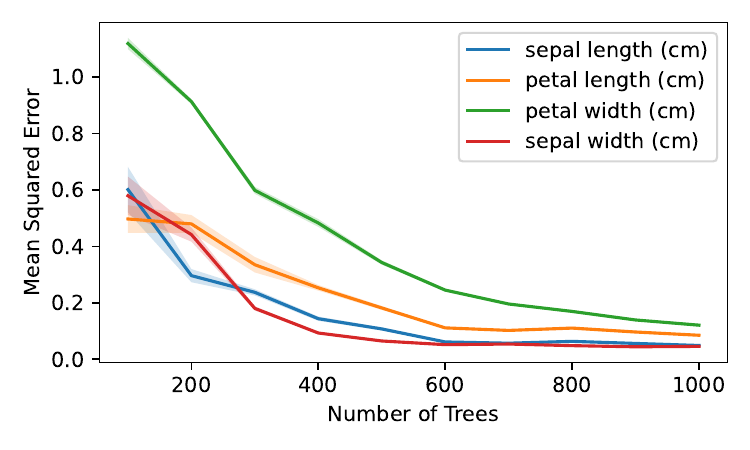}
    \subcaption{Iris dataset}
  \end{subfigure}
  \begin{subfigure}[b]{0.48\columnwidth}
    \centering
    \includegraphics[width=\textwidth]{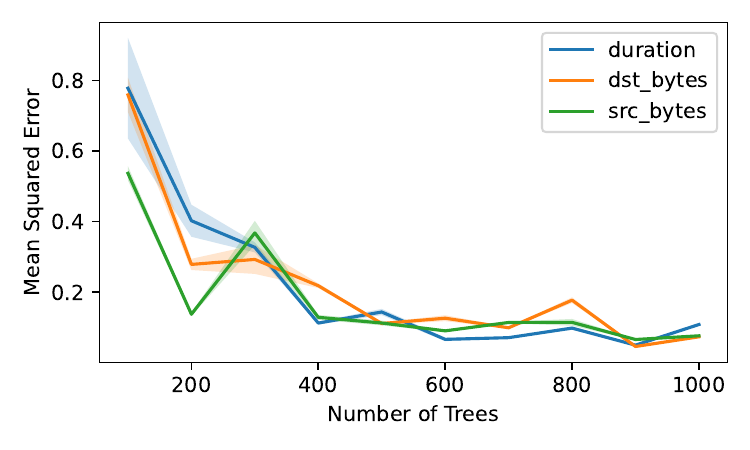}
    \subcaption{HTTP dataset}
  \end{subfigure}
  \hfill
  \begin{subfigure}[b]{0.48\columnwidth}
    \centering
    \includegraphics[width=\textwidth]{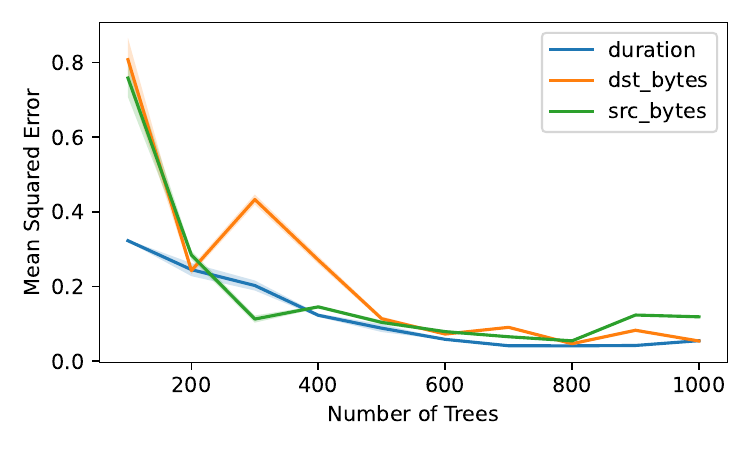}
    \subcaption{SMTP dataset}
  \end{subfigure}
  \caption{Mean-square errors of the real depths learned by iForest to the theoretical expected depths about numbers of trees. The shaded regions represent the confidence regions over multiple runs.}
  \label{fig:casestudy_empirical_studies_convergence}
\end{figure}

\begin{figure}
  \centering
  \begin{subfigure}[b]{0.93\columnwidth}
    \centering
    \includegraphics[width=\textwidth]{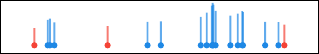}
    \caption{Example 1}
  \end{subfigure}
  \begin{subfigure}[b]{0.93\columnwidth}
    \centering
    \includegraphics[width=\textwidth]{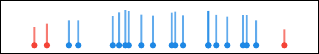}
    \caption{Example 2}
  \end{subfigure}
  \begin{subfigure}[b]{0.93\columnwidth}
    \centering
    \includegraphics[width=\textwidth]{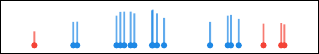}
    \caption{Example 3}
  \end{subfigure}
  \caption{Visualization of the expected depths for points generated from a uniform distribution. Circles and bars represent the point positions and the expected depths, respectively.}
  \label{fig:casestudy_empirical_studies_depth}
\end{figure}

\section{EMPIRICAL STUDIES} \label{section:empiricalstudies}
In this section, we conduct experiments to validate our theoretical findings. These include examining the convergence of the empirical depth and when iForest predicts a point as an anomaly.

\subsection{Convergence of the empirical depth}
This experiment is conducted using both synthetic and real-world datasets. We begin by independently sampling 100 points from three distributions: the standard normal distribution \( \mathcal{N}(0, 1) \), the uniform distribution \( U[0, 1] \), and the exponential distribution \( \textrm{Exp}(1) \). For real-world datasets, we examine the iris dataset~\citep{fisher36Iris}, which is a widely used machine learning dataset. Additionally, we examine the HTTP and SMTP datasets~\citep{kddcup99}, which are widely used anomaly detection datasets. Focusing on the one-dimensional case, we separately select each feature from all datasets to demonstrate the convergence of the empirical depth. For each dataset, we evaluate the mean squared errors of the depths learned by iForest to the theoretical expected depths with 100 subsamples as the number of iTrees ranges from 100 to 1000. Each setup is repeated 10 times independently. 

Figure~\ref{fig:casestudy_empirical_studies_convergence} shows the curves of the errors about the number of iTrees for different datasets. We can see that the error decreases rapidly as the number of trees increases, which implies that the depth of every point may eventually converge to the theoretically expected depth. This verifies the concentration property of the empirical depth to the theoretical depth and the correctness of our derived expected depth function shown in Proposition~\ref{proposition:concentration} and Theorem~\ref{theorem:expected_depth}, respectively.

\subsection{Anomaly detection of iForest}
We generate multiple datasets, each of which contains 20 points drawn from a uniform distribution independently. We compute the expected depth of each point to help understand when and when not iForest can detect anomalies. The results are presented in Figure~\ref{fig:casestudy_empirical_studies_depth}, where we label the points with a small depth considered as anomalies and a larger depth considered as normal points in red and blue colors, respectively.

From Figure~\ref{fig:casestudy_empirical_studies_depth}, three key observations can be made:
(1) Marginal single and clustered anomalies are correctly labeled, even though some are not significantly distant from normal points. This aligns with Theorems~\ref{theorem:marginal_single_anomaly} and~\ref{theorem:marginal_multiple_anomalies}, stating that marginal anomalies are more easily isolated.
(2) In Figure~\ref{fig:casestudy_empirical_studies_depth}~(a), only the central anomaly is detected. This is consistent with Theorem~\ref{theorem:central_single_anomaly}, which indicates that iForest can identify central anomalies only when they are sufficiently distant from normal data.
(3) In Figures~\ref{fig:casestudy_empirical_studies_depth}~(b) and~(c), $k$-NN may fail to detect the rightmost clustered anomaly when $k$ is set too small, whereas iForest successfully identifies all marginal clustered anomalies. This supports Theorems~\ref{theorem:marginal_multiple_anomalies}–\ref{theorem:marginal_multiple_anomalies_nearest_neighbor}, demonstrating that iForest is more adaptive to parameter settings than $k$-NN.

\section{DISCUSSIONS} \label{section:discussions}
Section~\ref{section:theory} provided theoretical studies of iForest in the one-dimensional case. In this section, we proceed to discuss the extended results of multi-dimensional cases.

\begin{figure}[t]
  \centering
  \includegraphics[width=0.465\textwidth]{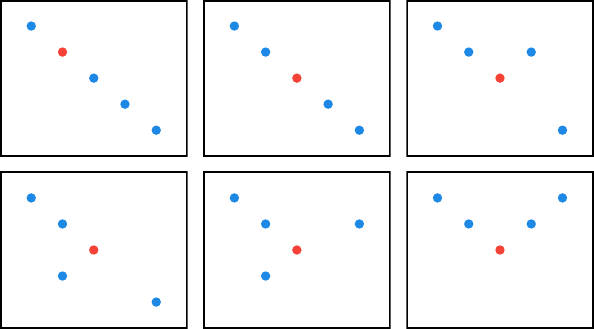}
  \caption{An example of six different points relative positions that lead to different forms of the expected depth function. Red points are the concerned points.}
  \label{fig:multiple_cases}
\end{figure}

\begin{figure}[t]
  \centering
  \includegraphics[width=0.465\textwidth]{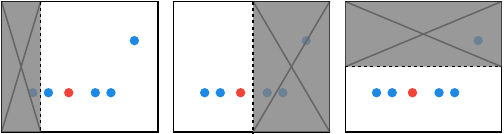}
  \caption{An example of a candidate split attribute and a split point changing after one split. Red points are the concerned points.}
  \label{fig:candidate_change}
\end{figure}

\textbf{Challenges in Multi-Dimensional Extensions} \vspace{0.5em} \\
In essence, there are two main challenges for theoretically analyzing the inductive bias of Isolation Trees in multi-dimensional cases. Firstly, there emerge a lot of relative positions between the concerned point and others for the multi-dimensional cases, which usually lead to various forms of the expected depth function. For example, in the simple two-dimensional cases with five points, there are at least six different forms of expected depth function for different point permutations, as shown in Figure~\ref{fig:multiple_cases}. Thus, it is difficult to employ a unified scheme to handle various expected depth functions. Secondly, the candidate splitting attribute and splitting point may completely change after one split due to the randomness of splittings. Taking Figure~\ref{fig:candidate_change} as an example, after one split, there are two candidate split attributes in the leftmost case but only one in the middle and right cases, which makes the growth process of Isolation Trees unpredictable.
By applying the same random walk model, just like the one-dimensional cases, the state space grows exponentially with the dimension $d$, making the analysis intractable. And there are almost no mature techniques for random processes to learn from~\citep{bhattacharya2021random}.

\textbf{A Negative Result in Two Dimensions} \vspace{0.5em} \\
To formally demonstrate the difficulty, we studied a two-dimensional toy example with four nearly collinear points. We have the following negative result:

\begin{proposition}
  For $D = \{\mathbf{x}_{1}, \mathbf{x}_{2}, \mathbf{x}_{3}, \mathbf{x}_{4}\}$ where $\mathbf{x}_{1} = (1, 1)$, $\mathbf{x}_{2} = (2, 3)$, $\mathbf{x}_{3} = (3, 4)$, and $\mathbf{x}_{4} = (4, 5)$, we have the followings:
  \begin{align*}
    \frac{745}{288} = \bar{h}(\mathbf{x}_{3}) & \neq \bar{h}(\mathbf{x}_{3}; \mathbf{x}_{1:3}) + \bar{h}(\mathbf{x}_{3}; \mathbf{x}_{3:4}) = \frac{744}{288} \ ,                                   \\
    \frac{491}{288} = \bar{h}(\mathbf{x}_{1}) & \neq \frac{\bar{h}(\mathbf{x}_{1}^{(1)}; \mathbf{x}_{1:4}^{(1)}) + \bar{h}(\mathbf{x}_{1}^{(2)}; \mathbf{x}_{1:4}^{(2)})}{2} = \frac{492}{288} \ ,
  \end{align*}
  where $\bar{h}(\mathbf{x}_{1}; \mathbf{x}_{1:4})$ and $\bar{h}(\mathbf{x}_{3}; \mathbf{x}_{1:4})$ are abbreviated as $\bar{h}(\mathbf{x}_{1})$ and $\bar{h}(\mathbf{x}_{3})$, respectively.
  \label{proposition:negative_result}
\end{proposition}

Proposition~\ref{proposition:negative_result} shows that even in such a simple two-dimensional setting, basic characteristics fail to hold:
(1) the expected depth cannot be decomposed into contributions from left and right subsets;
(2) it cannot be reduced to the average over the projections on all dimensions.
These violations indicate that closed-form expressions of the expected depth function may not exist in multi-dimensional settings and may require case-by-case analysis for different relative positions.

Although deriving exact formulas for multi-dimensional cases may be infeasible, it may be feasible to establish upper and lower bounds of normal and anomalous points in the purpose line of inductive bias analysis. Such analysis can be converted into identifying best- and worst-case scenarios, which are still tractable using the random walk framework, and it is interesting to explore this in the future.

In summary, while the theoretical analysis of iForest in multi-dimensional spaces remains challenging, our one-dimensional study provides essential foundations. The random walk model introduced here not only enables a precise characterization of the inductive bias in one dimension but also lays the groundwork for future explorations in multi-dimensional settings.

\section{CONCLUSIONS} \label{section:conclusions}

We theoretically investigated the inductive bias of iForest. Specifically, we introduced a random walk model for the growth process of iForest. This model facilitates the derivation of the closed-form expected depth function of iForest, allowing for a thorough comparison with the $k$-NN anomaly detector. Through case studies, we demonstrated that iForest exhibits less sensitivity in identifying central anomalies and greater parameter adaptability compared to $k$-NN. Numerical experiments verified our theoretical findings, including the concentration of iForest and the case studies.

\bibliography{bibliography.bib}

% \begin{thebibliography}{}
%   \setlength{\itemindent}{-\leftmargin}
%   \makeatletter\renewcommand{\@biblabel}[1]{}\makeatother
%   \bibitem{} J.~Alspector, B.~Gupta, and R.~B.~Allen (1989).
%   \newblock Performance of a stochastic learning microchip.
%   \newblock In D. S. Touretzky (ed.),
%   \textit{Advances in Neural Information Processing Systems 1}, 748--760.
%   San Mateo, Calif.: Morgan Kaufmann.

%   \bibitem{} F.~Rosenblatt (1962).
%   \newblock \textit{Principles of Neurodynamics.}
%   \newblock Washington, D.C.: Spartan Books.

%   \bibitem{} G.~Tesauro (1989).
%   \newblock Neurogammon wins computer Olympiad.
%   \newblock \textit{Neural Computation} \textbf{1}(3):321--323.
% \end{thebibliography}

%%%%%%%%%%%%%%%%%%%%%%%%%%%%%%%%%%%%%%%%%%%%%%%%%%%%%%%%%%%%

\section*{Checklist}

\begin{comment}
% %%% BEGIN INSTRUCTIONS %%%
The checklist follows the references. For each question, choose your answer from the three possible options: Yes, No, Not Applicable.  You are encouraged to include a justification to your answer, either by referencing the appropriate section of your paper or providing a brief inline description (1-2 sentences).
Please do not modify the questions.  Note that the Checklist section does not count towards the page limit. Not including the checklist in the first submission won't result in desk rejection, although in such case we will ask you to upload it during the author response period and include it in camera ready (if accepted).

\textbf{In your paper, please delete this instructions block and only keep the Checklist section heading above along with the questions/answers below.}
% %%% END INSTRUCTIONS %%%
\end{comment}

\begin{enumerate}

  \item For all models and algorithms presented, check if you include:
        \begin{enumerate}
          \item A clear description of the mathematical setting, assumptions, algorithm, and/or model. \\
          \textbf{Yes}, Sections~\ref{section:preliminary}-\ref{section:theory} provide the detailed description.
          \item An analysis of the properties and complexity (time, space, sample size) of any algorithm. \\
          \textbf{Not Applicable}, this paper is a theoretical paper studying existing algorithms.
          \item (Optional) Anonymized source code, with specification of all dependencies, including external libraries. \\
          \textbf{Yes}, please refer to the supplemental material.
        \end{enumerate}

  \item For any theoretical claim, check if you include:
        \begin{enumerate}
          \item Statements of the full set of assumptions of all theoretical results. \\
          \textbf{Yes}, Sections~\ref{section:preliminary}-\ref{section:casestudy} provide the detailed assumptions.
          \item Complete proofs of all theoretical results. \\
          \textbf{Yes}, please refer to the Appendix.
          \item Clear explanations of any assumptions. \\
          \textbf{Yes}, we have detailed explanations after all the assumptions.
        \end{enumerate}

  \item For all figures and tables that present empirical results, check if you include:
        \begin{enumerate}
          \item The code, data, and instructions needed to reproduce the main experimental results (either in the supplemental material or as a URL). \\
          \textbf{Yes}, please refer to the supplemental material.
          \item All the training details (e.g., data splits, hyperparameters, how they were chosen). \\
          \textbf{Yes}, please refer to the supplemental material.
          \item A clear definition of the specific measure or statistics and error bars (e.g., with respect to the random seed after running experiments multiple times). \\
          \textbf{Yes}, please refer to the supplemental material.
          \item A description of the computing infrastructure used. (e.g., type of GPUs, internal cluster, or cloud provider). \\
          \textbf{Yes}, please refer to the supplemental material.
        \end{enumerate}

  \item If you are using existing assets (e.g., code, data, models) or curating/releasing new assets, check if you include:
        \begin{enumerate}
          \item Citations of the creator If your work uses existing assets. \\
          \textbf{Yes}, we have cited the creator in the bibliography.
          \item The license information of the assets, if applicable. \\
          \textbf{Not Applicable}.
          \item New assets either in the supplemental material or as a URL, if applicable. \\
          \textbf{Not Applicable}.
          \item Information about consent from data providers/curators. \\
          \textbf{Not Applicable}.
          \item Discussion of sensible content if applicable, e.g., personally identifiable information or offensive content. \\
          \textbf{Not Applicable}.
        \end{enumerate}

  \item If you used crowdsourcing or conducted research with human subjects, check if you include:
        \begin{enumerate}
          \item The full text of instructions given to participants and screenshots. \\
          \textbf{Not Applicable}.
          \item Descriptions of potential participant risks, with links to Institutional Review Board (IRB) approvals if applicable. \\
          \textbf{Not Applicable}.
          \item The estimated hourly wage paid to participants and the total amount spent on participant compensation. \\
          \textbf{Not Applicable}.
        \end{enumerate}

\end{enumerate}

\clearpage
\appendix
\thispagestyle{empty}

% Supplementary material: To improve readability, you must use a single-column format for the supplementary material.
\onecolumn

\section{Proofs in Section~\ref{section:theory}}

Here, we provide the proofs of results in Sections~\ref{section:theory}.

% \subsection{Proof of Theorem~\ref{theorem:nofreelunch}}
% \label{appendix:proof_no_free_lunch}

% The proof of the ``No Free Lunch'' theorem for unsupervised anomaly detection is quite straightforward.
% As a learning algorithm only observes the input features, one can always construct a sequence of labels that is inconsistent with the algorithm output as follows.
% \[
%   y_{i} = \begin{cases}
%     1, & \text{if } \mathcal{A}(D)(\mathbf{x}_{i}) = 0 \ , \\
%     0, & \text{otherwise} \ ,
%   \end{cases}
% \]
% which always leads to a trivially high error.

\subsection{Proof of Proposition~\ref{proposition:concentration}}

We begin with the following technical lemma.

\begin{lemma}[Hoeffding's inequality~\citep{Hoeffding63Probability}] \label{lemma:hoeffding}
  Let $X_{1}, \dots, X_{n}$ be independent random variables such that $a_{i} \leq X_{i} \leq b_{i}$ for all $i \in [n]$.
  Consider the sum of these random variables $S_{n} = \frac{1}{n} \sum_{i = 1}^{n} X_{i}$.
  Then, for all $\epsilon > 0$, we have
  \begin{align*}
    \Pr \left[ \left| S - \mathbb{E}[S] \right| \geq \epsilon \right] \leq 2 \exp \left( - 2 \epsilon^{2}
    / \sum_{i = 1}^{n} (b_{i} - a_{i})^{2} \right) \ .
  \end{align*}
\end{lemma}

Then we prove Proposition~\ref{proposition:concentration}.

\textbf{Proof} \ \
We first denote \(h^{\prime}(\mathbf{x}; \mathbf{\Theta}_m) = M^{-1} h(\mathbf{x}; \mathbf{\Theta}_{m})\).
From the growth process of Isolation Trees, \( \mathbf{\Theta}_1, \dots, \mathbf{\Theta}_{m}\) are i.i.d. copies of \( \mathbf{\Theta} \), and we have
\[
  \mathbb{E}_{\mathbf{\Theta}} [ h^{\prime}(\mathbf{x}; \mathbf{\Theta}_m) ] = \mathbb{E}_{\mathbf{\Theta}} [ h^{\prime}(\mathbf{x}; \mathbf{\Theta})], ~ \forall 1 \leq m \leq M \ .
\]
By applying Hoeffding's inequality (Lemma~\ref{lemma:hoeffding}) to \(h^{\prime}(\mathbf{x}; \mathbf{\Theta}_m)\), we have
\begin{align*}
  \Pr \Bigg[
  \Bigg|
  \frac{1}{M} \sum_{m = 1}^{M} h(\mathbf{x}; D, \mathbf{\Theta}_{m}) -  \mathbb{E}_{\mathbf{\Theta}} [ h(\mathbf{x}; D, \mathbf{\Theta})] \Bigg| \geq \epsilon \Bigg] & = \Pr \left[ \left| \sum_{m = 1}^{M} h^{\prime}(\mathbf{x}; \mathbf{\Theta}_m) - \mathbb{E} \left[ \sum_{m = 1}^{M} h^{\prime}(\mathbf{x}; \mathbf{\Theta}_m) \right] \right| \geq \epsilon \right] \\
                                                                                                                                                                      & \leq 2 \exp \left( - 2 \epsilon^{2} / (M(n / M)^{2}) \right)                                                                                                                                        \\
                                                                                                                                                                      & = 2 \exp \left( - 2 \epsilon^{2} M / n^{2} \right) \ .
\end{align*}
This completes the proof.
$\hfill\square$

\subsection{Proof of Theorem~\ref{theorem:random_walk_model}}

\textbf{Markov property.} The Markov property holds directly from the growth process of Isolation Trees.

\textbf{Transition probability.} We now verify the transition probability.
Recall that the current state, the next state, and the absorbed state are denoted by \( (x_{\ell}, x_{r}) \) and \( (x_{\ell^{\prime}}, x_{r^{\prime}}) \), \( (x_{i}, x_{i}) \), respectively, where \( x_{\ell}, x_{\ell^{\prime}}, x_{r}, x_{r^{\prime}} , x_{i} \in D \), \( x_{\ell} \leq x_{i} \leq x_{r} \), and \( x_{\ell^{\prime}} \leq x_{i} \leq x_{r^{\prime}} \).
In every step, Isolation Tree chooses the split value in the interval \( (x_{\ell}, x_{i} ) \) or interval \( (x_{i}, x_{r}) \), and thus either \( x_{\ell^{\prime}} = x_{\ell} \) or \( x_{r^{\prime}} = x_{r} \) holds, implying that
\[
  \Pr \left[
    \mathbf{s}_{t + 1} = (x_{\ell^{\prime}}, x_{r^{\prime}}) \mid \mathbf{s}_{t} = (x_{\ell}, x_{r})
    \right] = 0 \ ,
\]
where \( x_{\ell^{\prime}}, x_{r^{\prime}} \in \{ (x, y) \mid (x \neq x_{\ell} \wedge y \neq x_{r}) \vee (x = x_{\ell} \wedge y \geq x_{r}) \vee (y = x_{r} \wedge x \leq x_{\ell}) \} \).

We now consider the case of \(x_{\ell^{\prime}} > x_{\ell}\) and \(x_{r^{\prime}} = x_{r}\).
The state change from \( (x_{\ell}, x_{r}) \) to \( (x_{\ell^{\prime}}, x_{r}) \) happens when the split point is chosen in the interval \( (x_{\ell^{\prime} - 1}, x_{\ell^{\prime}}) \). Hence, the transition probability equals the probability of choosing the split point in the interval \( (x_{\ell^{\prime} - 1}, x_{\ell^{\prime}}) \), which equals
\[
  \Pr \left[
    \mathbf{s}_{t + 1} = (x_{\ell^{\prime}}, x_{r}) \mid \mathbf{s}_{t} = (x_{\ell}, x_{r})
    \right] = \frac{x_{\ell^{\prime} - 1} - x_{\ell^{\prime}}}{x_{r} - x_{\ell}}, ~ \forall (x_{\ell} < x_{\ell^{\prime}} \leq x_{i}) \ .
\]
Similarly, for \( x_{\ell^{\prime}} = x_{\ell} \textrm{ and } x_{r^{\prime}} < x_{r} \), we have
\[
  \Pr \left[
    \mathbf{s}_{t + 1} = (x_{\ell}, x_{r^{\prime}}) \mid \mathbf{s}_{t} = (x_{\ell}, x_{r})
    \right] = \frac{x_{r^{\prime} + 1} - x_{r^{\prime}}}{x_{r} - x_{\ell}}, ~ \forall (x_{i} \leq x_{r^{\prime}} < x_{r}) \ .
\]
The proof is complete. $\hfill\square$

\subsection{Proof of Lemma~\ref{lemma:expected_depth}}

We will prove Lemma~\ref{lemma:expected_depth} first before proving Lemma~\ref{lemma:depth_decomposition}.

\textbf{Proof} \ \
By the definition of $\bar{h}$, it is equivalent to show that

\[
  \mathbb{E}_{\mathbf{\Theta}^{\prime}} \left[h \left( x_{i}; x^{\prime}, x_{1}, \dots, x_{i}, \mathbf{\Theta}^{\prime} \right)\right] = \frac{x_{1} - x^{\prime}}{x_{i} - x^{\prime}} + \mathbb{E}_{\mathbf{\Theta}} \left[ h \left( x_{i}; x_{1}, \dots, x_{i}, \mathbf{\Theta} \right) \right] \ .
\]

Let $\mathbf{\Theta}^{\prime} = (\Theta^{\prime}_{1}, \dots, \Theta^{\prime}_i)$ and $\mathbf{\Theta} = (\Theta_{1}, \dots, \Theta_{i - 1})$ be the vectors of random variables representing the split at each step after $x^{\prime}$ is added and before $x^{\prime}$ is added, respectively. Note that when $x_{i}$ is isolated, the randomness can be determined. Let $\Omega$ be the event that the first split is between $x^{\prime}$ and $x_{1}$. Then, we can expand the expectation as follows:

\[
  \begin{aligned}
    \mathbb{E}_{\mathbf{\Theta}^{\prime}} \Big[ h \big( x_{i}; x^{\prime}, x_{1}, \dots, x_{i}, \mathbf{\Theta}^{\prime} \big) \Big] = & \  \mathbb{E}_{\mathbf{\Theta}^{\prime}_{2: i}} \left[ \mathbb{E}_{\Theta^{\prime}_{1}} \left[ h \left( x_{i}; x^{\prime}, x_{1}, \dots, x_{i}, \mathbf{\Theta}^{\prime} \right) \right] \right]                                                                    \\
    =                                                                                                                                  & \  \Pr_{\Theta^{\prime}_{1}} \left[ \Theta^{\prime}_{1} \in \Omega \right] ~\mathbb{E}_{\mathbf{\Theta}^{\prime}_{2: i}} \left[ h \left( x_{i}; x^{\prime}, x_{1}, \dots, x_{i}, \mathbf{\Theta}^{\prime} \right) \mid \Theta^{\prime}_{1} \in \Omega \right] +     \\
                                                                                                                                       & \  \Pr_{\Theta^{\prime}_{1}} \left[ \Theta^{\prime}_{1} \notin \Omega \right] ~\mathbb{E}_{\mathbf{\Theta}^{\prime}_{2: i}} \left[ h \left( x_{i}; x^{\prime}, x_{1}, \dots, x_{i}, \mathbf{\Theta}^{\prime} \right) \mid \Theta^{\prime}_{1} \notin \Omega \right] \\
    =                                                                                                                                  & \  \frac{x_{1} - x^{\prime}}{x_{i} - x^{\prime}} ~\left(1 + \mathbb{E}_{\mathbf{\Theta}^{\prime}_{2: i}} \left[ h \left( x_{i}; x^{\prime}, x_{1}, \dots, x_{i}, \mathbf{\Theta}^{\prime} \right) \mid \Theta^{\prime}_{1} \in \Omega \right] \right) +             \\
                                                                                                                                       & \  \frac{x_{i} - x_{1}}{x_{i} - x^{\prime}} ~\left(1 + \mathbb{E}_{\mathbf{\Theta}^{\prime}_{2: i}} \left[ h \left( x_{i}; x^{\prime}, x_{1}, \dots, x_{i}, \mathbf{\Theta}^{\prime} \right) \mid \Theta^{\prime}_{1} \notin \Omega \right] \right) \ .
  \end{aligned}
\]

For $\Theta_{1}^{\prime} \in \Omega$, the tree growth process after the first split is the same as the case where $x^{\prime}$ is not added. Thus, we have

\[
  \mathbb{E}_{\mathbf{\Theta}^{\prime}_{2: i}} \left[ h \left( x_{i}; x^{\prime}, x_{1}, \dots, x_{i}, \mathbf{\Theta}^{\prime} \right) \mid \Theta^{\prime}_{1} \in \Omega \right] = \mathbb{E}_{\mathbf{\Theta}_{1: i - 1}} \left[ h \left( x_{i}; x_{1}, \dots, x_{i}, \mathbf{\Theta} \right) \right] \ .
\]

For $\Theta_{1} \notin \Omega$, the tree growth process is the same as the case where $x^{\prime}$ is added. Thus, we have

\[
  \mathbb{E}_{\mathbf{\Theta}^{\prime}_{2: i}} \left[ h \left( x_{i}; x^{\prime}, x_{1}, \dots, x_{i}, \mathbf{\Theta}^{\prime} \right) \mid \Theta^{\prime}_{1} \notin \Omega \right] = \mathbb{E}_{\mathbf{\Theta}_{1: i - 1}} \left[ h \left( x_{i}; x_{1}, \dots, x_{i}, \mathbf{\Theta} \right) \right] - 1 \ .
\]

Therefore, we have

\[
  \begin{aligned}
    \mathbb{E}_{\mathbf{\Theta}^{\prime}} \left[ h \left( x_{i}; x^{\prime}, x_{1}, \dots, x_{i}, \mathbf{\Theta}^{\prime} \right) \right] = & \  \frac{x_{1} - x^{\prime}}{x_{i} - x^{\prime}} ~\left(1 + \mathbb{E}_{\mathbf{\Theta}^{\prime}_{2: i}} \left[ h \left( x_{i}; x^{\prime}, x_{1}, \dots, x_{i}, \mathbf{\Theta}^{\prime} \right) \mid \Theta^{\prime}_{1} \in \Omega \right] \right) + \\
                                                                                                                                             & \  \frac{x_{i} - x_{1}}{x_{i} - x^{\prime}} ~\left(1 + \mathbb{E}_{\mathbf{\Theta}^{\prime}_{2: i}} \left[ h \left( x_{i}; x^{\prime}, x_{1}, \dots, x_{i}, \mathbf{\Theta}^{\prime} \right) \mid \Theta^{\prime}_{1} \notin \Omega \right] \right)     \\
    =                                                                                                                                        & \  \frac{x_{1} - x^{\prime}}{x_{i} - x^{\prime}} + \mathbb{E}_{\mathbf{\Theta}} \left[ h \left( x_{i}; x_{1}, \dots, x_{i}, \mathbf{\Theta} \right) \right] \ .
  \end{aligned}
\]

The proof is complete.
$\hfill\square$

\subsection{Proof of Lemma~\ref{lemma:depth_decomposition}}

The verify
We begin with the following lemma.
\begin{lemma} \label{lemma:add_point_depth}
  For any given points \( x_{1} < x_{2} < \dots < x_{i} \) and \( x^{\prime} < x_{1} \), we have for all \( 1 \leq i \leq n \)
  \[
    \bar{h} \left( x_{i}; x^{\prime}, x_{1}, \dots, x_{n} \right) =
    \frac{x_{1} - x^{\prime}}{x_{i} - x^{\prime}} + \bar{h} \left( x_{i}; x_{1}, \dots, x_{n} \right) \ .
  \]
\end{lemma}

\textbf{Proof} \ \
Assume that $n > i$, otherwise it recovers Lemma~\ref{lemma:expected_depth}. By the definition of $\bar{h}$, we are to show that

\begin{equation}
  \mathbb{E}_{\mathbf{\Theta}^{\prime}} \left[ h \left( x_{i}; x^{\prime}, x_{1}, \dots, x_{n}, \mathbf{\Theta}^{\prime} \right) \right] = \frac{x_{1} - x^{\prime}}{x_{i} - x^{\prime}} + \mathbb{E}_{\mathbf{\Theta}} \left[ h \left( x_{i}; x_{1}, \dots, x_{n}, \mathbf{\Theta} \right) \right] \ . \label{equation:add_point_depth}
\end{equation}

Applying mathematical induction on $n$. For $n = i + 1$, we have the base case holds as follows:

\[
  \begin{aligned}
    \mathbb{E}_{\mathbf{\Theta}^{\prime}} \left[ h \left( x_{i}; x^{\prime}, x_{1}, \dots, x_{i + 1}, \mathbf{\Theta}^{\prime} \right) \right] & = \mathbb{E}_{\mathbf{\Theta}^{\prime}} \left[ h \left( x_{i}; x^{\prime}, x_{1}, \dots, x_{i}, \mathbf{\Theta}^{\prime} \right) \right] + 1                       \\
                                                                                                                                               & = \frac{x_{1} - x^{\prime}}{x_{i} - x^{\prime}} + \mathbb{E}_{\mathbf{\Theta}} \left[ h \left( x_{i}; x_{1}, \dots, x_{i}, \mathbf{\Theta} \right) \right] + 1     \\
                                                                                                                                               & = \frac{x_{1} - x^{\prime}}{x_{i} - x^{\prime}} + \mathbb{E}_{\mathbf{\Theta}} \left[ h \left( x_{i}; x_{1}, \dots, x_{i + 1}, \mathbf{\Theta} \right) \right] \ ,
  \end{aligned}
\]

where the first equality follows from the fact that the new point $x_{i + 1}$ increases the depth of $x_{i}$ by 1, and the second equality follows from Lemma~\ref{lemma:expected_depth}.

Assuming that Eq.~\eqref{equation:add_point_depth} holds for all $n < k \  (k > i + 1)$, we then consider $n = k$. There are three cases: If the first split occurs in $[x^{\prime}, x_{1}]$, the expected depth of $x_{i}$ increases by 1; If the first split occurs in $[x_{1}, x_{i}]$, the expected depth of $x_{i}$ remains unchanged; otherwise, a split in $[x^{\prime}, x_{i}]$ will eventually occur, and the expected depth of $x_{i}$ increases by about $\frac{x_{1} - x^{\prime}}{x_{i} - x^{\prime}}$. Formally, we have

\[
  \begin{aligned}
    \mathbb{E}_{\mathbf{\Theta}^{\prime}} \left[ h \left( x_{i}; x^{\prime}, x_{1}, \dots, x_{k}, \mathbf{\Theta}^{\prime} \right) \right] & = \frac{x_{1} - x^{\prime}}{x_{k} - x^{\prime}}(1 + \mathbb{E}_{\mathbf{\Theta}} \left[ h \left( x_{i}; x_{1}, \dots, x_{k}, \mathbf{\Theta} \right) \right])                                                                 \\
                                                                                                                                           & \quad\quad + \sum_{j = 2}^{i} \frac{x_{j} - x_{j - 1}}{x_{k} - x^{\prime}}\mathbb{E}_{\mathbf{\Theta}} \left[ h \left( x_{i};x_{1}, \dots, x_{k}, \mathbf{\Theta} \right) \right]                                             \\
                                                                                                                                           & \quad\quad + \sum_{j = i + 1}^{k} \frac{x_{j} - x_{j - 1}}{x_{k} - x^{\prime}} \mathbb{E}_{\mathbf{\Theta}^{\prime}} \left[ h \left( x_{i}; x^{\prime}, x_{1}, \dots, x_{j - 1}, \mathbf{\Theta}^{\prime} \right) \right] \ .
  \end{aligned}
\]

By induction hypothesis, we have

\[
  \mathbb{E}_{\mathbf{\Theta}^{\prime}} \left[ h \left( x_{i}; x^{\prime}, x_{1}, \dots, x_{j - 1}, \mathbf{\Theta}^{\prime} \right) \right] = \frac{x_{1} - x^{\prime}}{x_{i} - x^{\prime}} + \mathbb{E}_{\mathbf{\Theta}} \left[ h \left( x_{i}; x_{1}, \dots, x_{j - 1}, \mathbf{\Theta} \right) \right] \ ,
\]

which implies that

\begin{align*}
  \sum_{j = i + 1}^{k} \frac{x_{j} - x_{j - 1}}{x_{k} - x^{\prime}} \mathbb{E}_{\mathbf{\Theta}^{\prime}} & \left[ h \left( x_{i}; x^{\prime}, x_{1}, \dots, x_{j - 1}, \mathbf{\Theta}^{\prime} \right) \right]                                                                                                                         \\
                                                                                                          & = \frac{x_{k} - x_{i}}{x_{k} - x^{\prime}} \cdot \left( \frac{x_{1} - x^{\prime}}{x_{i} - x^{\prime}} + \mathbb{E}_{\mathbf{\Theta}} \left[ h \left( x_{i}; x_{1}, \dots, x_{k}, \mathbf{\Theta} \right) \right] \right) \ .
\end{align*}

Therefore, we conclude that

\[
  \begin{aligned}
    \mathbb{E}_{\mathbf{\Theta}^{\prime}} \left[ h \left( x_{i}; x^{\prime}, x_{1}, \dots, x_{k}, \mathbf{\Theta}^{\prime} \right) \right] = & \ \frac{x_{1} - x^{\prime}}{x_{k} - x^{\prime}} + \frac{x_{k} - x_{i}}{x_{k} - x^{\prime}} \cdot \frac{x_{1} - x^{\prime}}{x_{i} - x^{\prime}} +                                                                                                               \\
                                                                                                                                             & \  \left( \frac{x_{1} - x^{\prime}}{x_{k} - x^{\prime}} + \frac{x_{i} - x_{1}}{x_{k} - x^{\prime}} + \frac{x_{k} - x_{i}}{x_{k} - x^{\prime}} \right) \mathbb{E}_{\mathbf{\Theta}} \left[ h \left( x_{i}; x_{1}, \dots, x_{k}, \mathbf{\Theta} \right) \right] \\
    =                                                                                                                                        & \ \frac{x_{1} - x^{\prime}}{x_{i} - x^{\prime}} + \mathbb{E}_{\mathbf{\Theta}} \left[ h \left( x_{i}; x_{1}, \dots, x_{k}, \mathbf{\Theta} \right) \right] \ .
  \end{aligned}
\]

The proof is complete. $\hfill\square$

Then, we prove Lemma~\ref{lemma:depth_decomposition}. \\
\textbf{Proof} \ \
By Lemma~\ref{lemma:add_point_depth}, we have
\begin{align*}
  \bar{h} \left( x_{i}; x_{1}, \dots, x_{n} \right)
   & = \bar{h} \left( x_{i}; x_{1}, \dots, x_{n - 1} \right) + \frac{x_{n} - x_{n - 1}}{x_{n} - x_{i}}                                                   \\
   & = \bar{h} \left( x_{i}; x_{1}, \dots, x_{n - 2} \right) + \frac{x_{n - 1} - x_{n - 2}}{x_{n - 1} - x_{i}} + \frac{x_{n} - x_{n - 1}}{x_{n} - x_{i}} \\
   & = \bar{h} \left( x_{i}; x_{1}, \dots, x_{i} \right) + \frac{x_{i + 1} - x_{i}}{x_{i + 1} - x_{i}} + \dots + \frac{x_{n} - x_{n - 1}}{x_{n} - x_{i}} \\
   & = \bar{h} \left( x_{i}; x_{1}, \dots, x_{i} \right) + \bar{h} \left( x_{i}; x_{i}, \dots, x_{n} \right) \ ,
\end{align*}
which completes the proof.
$\hfill\square$

\subsection{Depth Function of Points Not in the Dataset}
For data points not in the dataset, we have the following theorem.
\begin{theorem}
  \label{theorem:expected_depth_outer}
  For any given dataset $D$ with sample size $n > 2$ and $x \notin D$, we have
  \begin{equation*}
    \bar{h}(x)= \left\{ \begin{aligned}
      \bar{h}(x_{1}) \ , & \quad \text{if } x < x_{1} \ ,                \\
      \bar{h}(x_{n}) \ , & \quad \text{if } x \geq x_{n} \ ,             \\
      \ell_{i}(x) \ ,    & \quad \text{if } x_{i} \leq x < x_{i + 1} \ ,
    \end{aligned} \right.
  \end{equation*}
  where
  \[
    \ell_{i}(x) = \bar{h}(x_{i}) + \frac{x - x_{i}}{x_{i + 1} - x_{i}} \left( \bar{h}(x_{i + 1}) - \bar{h}(x_{i}) \right)
  \]
  is the linear interpolation of $\bar{h}(x_{i})$ and $\bar{h}(x_{i + 1})$.
\end{theorem}

\textbf{Proof} \ \ For \( x < x_{1} \) or \( x \geq x_{n} \), evidently, isolating \( x \) is equivalent to isolating \( x_{1} \) or \( x_{n} \), respectively. Thus, we have
\[
  \mathbb{E}_{\mathbf{\Theta}} \left[
    h \left( x; x_{1}, \dots, x_{n}, \mathbf{\Theta} \right)
    \right]
  =
  \mathbb{E}_{\mathbf{\Theta}} \left[
    h \left( x_{1}; x_{1}, \dots, x_{n}, \mathbf{\Theta} \right)
    \right], \textrm{for } \forall x < x_{1}
\]
and
\[
  \mathbb{E}_{\mathbf{\Theta}} \left[
    h \left( x; x_{1}, \dots, x_{n}, \mathbf{\Theta} \right)
    \right]
  =
  \mathbb{E}_{\mathbf{\Theta}} \left[
    h \left( x_{n}; x_{1}, \dots, x_{n}, \mathbf{\Theta} \right)
    \right], \textrm{for } \forall x \geq x_{n} \ .
\]
We now consider the case where \( x \in \left[ x_{1}, x_{n} \right) \).
By Lemma~\ref{lemma:expected_depth}, it suffices to show that for \( x \in \left[ x_{i}, x_{i + 1} \right) \), the expected depth \( \mathbb{E}_{\mathbf{\Theta}} \left[ h \left( x; x_{1}, \dots, x_{n}, \mathbf{\Theta} \right) \right] \) is a linear function.
For any given \( \mathbf{\Theta} = \mathbf{\Theta}_{0} \), we have
\[
  h \left( x; x_{1}, \dots, x_{n}, \mathbf{\Theta}_{0} \right)
\]
is piecewise-constant.
Note that every split that does not isolation \( x \) may not change the value of \( h \left( x; x_{1}, \dots, x_{n}, \mathbf{\Theta}_{0} \right) \).
Therefore, we will focus on the split that isolates \( x \).
Denote by \( \Omega \) be the event that \( x \) is eventually isolated by a split in \( (x_{i}, x) \).
Then, we have
\begin{align*}
  \mathbb{E}_{\mathbf{\Theta}} \left[
    h \left( x; x_{1}, \dots, x_{n}, \mathbf{\Theta} \right)
    \right]
  =
   & \ \Pr[\mathbf{\Theta} \in \Omega] ~
  \mathbb{E}_{\mathbf{\Theta}} \left[
    h \left( x; x_{1}, \dots, x_{n}, \mathbf{\Theta} \right)
    \mid \Theta \in \Omega
  \right]                                   \\
   & + \Pr[\mathbf{\Theta} \notin \Omega] ~
  \mathbb{E}_{\mathbf{\Theta}} \left[
    h \left( x; x_{1}, \dots, x_{n}, \mathbf{\Theta} \right)
    \mid \Theta \notin \Omega
    \right] \ .
\end{align*}
When the split isolating \( x \) is in interval \( (x_{i}, x) \), point \( x \) is treated the same as \( x_{i + 1} \) and thus has the same depth as \( x_{i + 1} \).
Similarly, when the split isolating \( x \) is in interval \( (x, x_{i + 1}) \), point \( x \) is treated the same as \( x_{i} \) and thus has the same depth as \( x_{i} \).
Therefore, we have the depth of \( x \) is a constant given either \( \mathbf{\Theta} \in \Omega \) or \( \mathbf{\Theta} \notin \Omega \), which implies that
\[
  \mathbb{E}_{\mathbf{\Theta}} \left[
    h \left( x; x_{1}, \dots, x_{n}, \mathbf{\Theta} \right)
    \right]
  =
  \Pr[\mathbf{\Theta} \in \Omega] c_{1} + \Pr[\mathbf{\Theta} \notin \Omega] c_{2} \ ,
\]
where \( c_{1} \) and \( c_{2} \) are constants.
Then it suffices to analyze the probability of event \( \Omega \).
By the growth mechanism of Isolation Forest, the split that isolates \( x \) is uniformly sampled from the interval \( (x_{i}, x_{i + 1}) \), and thus we have
\[
  \Pr[\mathbf{\Theta} \in \Omega] = \frac{x - x_{i}}{x_{i + 1} - x_{i}}
  \quad \textrm{and} \quad
  \Pr[\mathbf{\Theta} \notin \Omega] = \frac{x_{i + 1} - x}{x_{i + 1} - x_{i}} \ .
\]
As the dataset is fixed, \( \Pr[\mathbf{\Theta} \in \Omega] \) and \( \Pr[\mathbf{\Theta} \notin \Omega] \) are both linear in \( x \), which completes the proof.

$\hfill\square$

\section{Further Elaboration of Assumption~\ref{assumption:density_ratio}}
\label{section:elaboration_assumption_kappa}

Recall that we assumed \( \kappa > \Omega(\sqrt{n + 3}) \) in Assumption~\ref{assumption:density_ratio}.
Here, we will show that \( \kappa > \Omega(\sqrt{n}) \) is commonly satisfied in practice.
Note that any distribution can be decomposed into a mixture of uniform distributions, and we will focus on the uniform distribution, for which we have the following proposition.
\begin{proposition}
  \label{proposition:uniform_kappa}
  Let \( X_{1}, \dots, X_{n}, n > 3 \) be i.i.d.\ random variables from \( \mathcal{U}[0, 1] \).
  Then, with probability \( 1 - O(1 / n^{1/2}) \), we have the following holds
  \[
    \kappa \geq \frac{1}{2} \sqrt{n} \ .
  \]
\end{proposition}
\textbf{Proof} \ \
Let \( L = \min_{i} ~ \lvert X_{i + 1} - X_{i} \rvert \).
We first show that
\[
  \mathbb{E}[L] = 1 / (n^{2} - 1) \ .
\]
By the symmetry of order statistics, we have
\[
  \Pr[L > t] = n! \Pr[L > t, X_{1} < \dots < X_{n}] \ .
\]
Observing that \( \Pr[L > t, X_{1} < \dots < X_{n}] \) equals the volume of the set
\[
  S = \{ (x_{1}, \dots, x_{n}) \in [0, 1]^{n} \mid x_{i} + t \leq x_{i + 1}, i = 1, \dots, n - 1 \} \ .
\]
By applying the transformation
\[
  (y_{1}, \dots, y_{n}) = (x_{1}, x_{2} - t, x_{3} - 2t, \dots, x_{n} - (n - 1)t) \ ,
\]
which is a volume-preserving transformation, we have a new set
\[
  S^{\prime} = \{ (x_{1}, \dots, x_{n}) \in [0, 1 - (n - 1)t]^{n} \mid x_{i} + t \leq x_{i + 1}, i = 1, \dots, n - 1 \} \ .
\]
Again by the symmetry of order statistics, we have
\begin{align*}
  \Pr[L > l, X_{1} < \dots < X_{n}] & = n! ~ \textrm{Vol}(S^{\prime})                                 \\
                                    & = n! ~ \frac{1}{n!} (1 - (n - 1)t)^{n} = (1 - (n - 1)t)^{n} \ .
\end{align*}
Therefore, we have
\begin{align*}
  \mathbb{E}[L] = \int_{0}^{1 / (n - 1)} \Pr[L > t] \, dt = \int_{0}^{1 / (n - 1)} (1 - (n - 1)t)^{n} \, dt = \frac{1}{n^{2} - 1} \ .
\end{align*}
By Markov's inequality, we have
\[
  \Pr \left[ L > \frac{1}{n^{2} - 1} + \frac{1}{n^{3/2}} \right] \leq \frac{n^{3 / 2}}{n^{2} - 1} = O(1 / n^{1 / 2}) \ .
\]
Observing that
\[
  \Pr[\max_{i} X_{i} - \min_{i} X_{i} \leq 1 / 2] \leq \frac{c}{2^n} \ ,
\]
for some \( c > 0 \), implying that
\[
  \Pr \left[ U \leq \frac{1}{2n} \right] \leq \frac{c}{2^n} \ ,
\]
Therefore, we have
\[
  \Pr \left[ \frac{U}{L} \leq \frac{\frac{1}{2n}}{\frac{1}{(n^{2} - 1)} + \frac{1}{n^{3 / 2}}} \right] \leq O(n^{-\frac{1}{2}} + 2^{-n}) = O(n^{-\frac{1}{2}}) \ ,
\]
which completes the proof. \(\hfill\square\)

\section{Proofs of Theorems~\ref{theorem:marginal_single_anomaly}-\ref{theorem:marginal_single_anomaly_nearest_neighbor}}
We will prove Theorems about marginal single anomaly here, including Theorem~\ref{theorem:marginal_single_anomaly}-\ref{theorem:marginal_single_anomaly_nearest_neighbor}.

\subsection{Proof of Theorem~\ref{theorem:marginal_single_anomaly}}
\label{appendix:proof_marginal_single_anomaly}

To avoid ambiguity, we denote by \( U_{\textrm{ms}} = U \) and \( L_{\textrm{ms}} = L \) following the definitions in Definition~\ref{definition:density_metrics}.
Suppose that \( x_{2} - x_{1} > U_{\textrm{ms}} \cdot \kappa \).
We will prove by showing that for all \( j > 1 \), the following holds
\[
  \bar{h}(x_{1}; x_{1: n}) <
  \sup_{x_{1: n}} \bar{h}(x_{1}; x_{1: n}) \leq
  \inf_{j > 1, x_{1: n}} \bar{h}(x_{j}; x_{1: n}) \leq
  \bar{h}(x_{j}; x_{1: n}) \ ,
\]
where we take \( \sup \) and \( \inf \) instead of \( \max \) and \( \min \) because the maximal and the minimal may not exist when
\begin{equation*}
  x_{2} - x_{1} > U_{\textrm{ms}} \cdot \kappa \ .
\end{equation*}
Note that
\[
  \bar{h}(x_{1}; x_{1: n}) < \sup_{x_{1: n}} \bar{h}(x_{1}; x_{1: n}) \quad \textrm{and} \quad \inf_{j > 1, x_{1: n}} \bar{h}(x_{j}; x_{1: n}) \leq \bar{h}(x_{j}; x_{1: n})
\]
are trivial.
Then it suffices to show that
\[ \sup_{x_{1: n}} \bar{h}(x_{1}; x_{1: n}) \leq \inf_{j > 1, x_{1: n}} \bar{h}(x_{j}; x_{1: n}) \ . \]
For convenience, we define
\[
  s_{i} \triangleq x_{i + 1} - x_{i}, i \leq n - 1 \ .
\]
Apparently, assignment to \( (x_{1}, \dots, x_{n}) \) is equivalent to assigning \( (s_{1}, \dots, s_{n - 1}) \).
From the condition \( U_{\textrm{ms}} = \max_{i \geq 2} ~ \lvert x_{i + 1} - x_{i} \rvert \) and \( x_{2} - x_{1} > U_{\textrm{ms}} \cdot \kappa \), we have
\[
  s_{1} > U_{\textrm{ms}} \cdot \kappa \geq U_{\textrm{ms}} \geq s_{j}, ~ \forall j > 1 \ .
\]
We assert that \( \inf_{j > 1, x_{1: n}} \bar{h}(x_{j}; x_{1: n}) \) is achieved at \( j = n \), i.e.,
\[
  \inf_{j > 1, x_{1: n}} \bar{h}(x_{j}; x_{1: n}) = \inf_{x_{1: n}} \bar{h}(x_{n}; x_{1: n}) \ .
\]
Otherwise, there exists a \( j_{0} \) satisfying \( 1 < j_{0} < n \) such that
\[
  \inf_{x_{1: n}} \bar{h}(x_{j_{0}}; x_{1: n}) = \inf_{j > 1, x_{1: n}} \bar{h}(x_{j}; x_{1: n}) < \inf_{x_{1: n}} \bar{h}(x_{n}; x_{1: n}) \ .
\]
By Lemma~\ref{lemma:expected_depth}, we have
\begin{align*}
  \bar{h}(x_{j_{0}}; x_{1: n}) =
   & ~ \frac{s_{1}}{s_{1} + \dots + s_{j_{0} - 1}} + \frac{s_{2}}{s_{2} + \dots + s_{j_{0} - 1}} + \dots + \frac{s_{j_{0} - 1}}{s_{j_{0} - 1}}       \\
   & ~ + \frac{s_{j_{0}}}{s_{j_{0}}} + \frac{s_{j_{0} + 1}}{s_{j_{0}} + s_{j_{0} + 1}} + \dots + \frac{s_{n - 1}}{s_{j_{0}} + \dots + s_{n - 1}} \ .
\end{align*}
Notice that
\begin{align*}
  \frac{s_{j_{0}}}{s_{j_{0}}} +
  \frac{s_{j_{0} + 1}}{s_{j_{0}} + s_{j_{0} + 1}} & +
  \dots +
  \frac{s_{n - 1}}{s_{j_{0}} + \dots + s_{n - 1}}
  \\
  >
                                                  & \ \frac{s_{j_{0}}}{s_{j_{0}} + s_{1} + \dots + s_{j_{0} - 1}} +
  \frac{s_{j_{0} + 1}}{s_{j_{0} + 1} + s_{j_{0}} + s_{1} + \dots + s_{j_{0} - 1}}                                                             \\
                                                  & + \dots + \frac{s_{n - 1}}{s_{n - 1} + \dots + s_{j_{0}} + s_{1} + \dots + s_{j_{0} - 1}}
  \\
  =
                                                  & \ \sum_{i = j_{0}}^{n - 1} {
  \frac{s_{i}}{s_{i} + \dots + s_{j_{0}} + s_{1} + \dots + s_{j_{0} - 1}}
  } \ .
\end{align*}
Therefore, we have
\[
  \bar{h}(x_{j_{0}}; x_{1: n}) >
  \sum_{i = 1}^{j_{0} - 1} {
  \frac{s_{i}}{s_{i} + \dots + s_{j_{0} - 1}}
  }
  +
  \sum_{i = j_{0}}^{n - 1} {
  \frac{s_{i}}{s_{i} + \dots + s_{j_{0}} + s_{1} + \dots + s_{j_{0} - 1}}
  } \ .
\]
If we re-assign \( (s^{\prime}_{1}, \dots, s^{\prime}_{n - 1}) \) with \( (s_{n - 1}, \dots, s_{j_{0}}, s_{1}, \dots, s_{j_{0} - 1}) \) and re-assign the corresponding \( (x^{\prime}_{1}, \dots, x^{\prime}_{n}) \) with \( s_{i}^{\prime} \).
Then, we have
\[
  \bar{h}(x^{\prime}_{n}; x^{\prime}_{1: n}) =
  \sum_{i = 1}^{j_{0} - 1} {
  \frac{s_{i}}{s_{i} + \dots + s_{j_{0} - 1}}
  }
  +
  \sum_{i = j_{0}}^{n - 1} {
  \frac{s_{i}}{s_{i} + \dots + s_{j_{0}} + s_{1} + \dots + s_{j_{0} - 1}}
  } <
  \bar{h}(x_{j_{0}}; x_{1: n}) \ ,
\]
which conflicts with the condition that \( \bar{h}(x_{j_{0}}; x_{1: n}) \) is the minimal.
To this end, it suffices to show that
\[
  \sup_{x_{1: n}} \bar{h}(x_{1}; x_{1: n}) < \inf_{x_{1: n}} \bar{h}(x_{n}; x_{1: n}) \ .
\]
Recall that
\[
  \bar{h}(x_{1}; x_{1: n}) = \frac{s_{1}}{s_{1}} + \frac{s_{2}}{s_{1} + s_{2}} + \dots + \frac{s_{n - 1}}{s_{1} + \dots + s_{n - 1}} \triangleq \tilde{h}(x_{1}; s_{1}, \dots, s_{n - 1}) \ ,
\]
which is decreasing with respect to \( s_{1} \) and increasing with respect to \( s_{n - 1} \), implying that
\begin{align*}
    & \ \sup_{x_{1: n}} \bar{h}(x_{1}; x_{1: n})                                                                                                                                                                                                                               \\
  = & \ \sup_{s_{2}, \dots, s_{n - 2}} \tilde{h}(x_{1}; U_{\textrm{ms}} \cdot \kappa, s_{2}, \dots, s_{n - 2}, U_{\textrm{ms}})                                                                                                                                                \\
  = & \ \sup_{s_{2}, \dots, s_{n - 2}} \frac{U_{\textrm{ms}} \cdot \kappa}{U_{\textrm{ms}} \cdot \kappa} + \frac{s_{2}}{U_{\textrm{ms}} \cdot \kappa + s_{2}} + \dots + \frac{U_{\textrm{ms}}}{U_{\textrm{ms}} \cdot \kappa + s_{2} + \dots + s_{n - 2} + U_{\textrm{ms}}} \ .
\end{align*}
We now consider the following optimization problem
\begin{align*}
  \max_{s_{2}, \dots, s_{n - 2}} & \quad \frac{U_{\textrm{ms}} \cdot \kappa}{U_{\textrm{ms}} \cdot \kappa} + \frac{s_{2}}{U_{\textrm{ms}} \cdot \kappa + s_{2}} + \dots + \frac{U_{\textrm{ms}}}{U_{\textrm{ms}} \cdot \kappa + s_{2} + \dots + s_{n - 2} + U_{\textrm{ms}}} \ , \\
  \textrm{s.t. }                 & \quad L_{\textrm{ms}} \leq s_{i} \leq U_{\textrm{ms}}, \quad \forall i \ ,
\end{align*}
which has a Lagrangian function as follows:
\begin{align*}
  L(s_{2}, \dots, s_{n - 2}) = & \ \frac{U_{\textrm{ms}} \cdot \kappa}{U_{\textrm{ms}} \cdot \kappa} + \frac{s_{2}}{U_{\textrm{ms}} \cdot \kappa + s_{2}} + \dots + \frac{U_{\textrm{ms}}}{U_{\textrm{ms}} \cdot \kappa + s_{2} + \dots + s_{n - 2} + U_{\textrm{ms}}} \\
                               & \ + \mu_{2} \left( L_{\textrm{ms}} - s_{2} \right) + \dots + \mu_{n - 2} \left( L_{\textrm{ms}} - s_{n - 2} \right)                                                                                                                   \\
                               & \ + \mu_{2}^{\prime} \left( s_{2} - U_{\textrm{ms}} \right) + \dots + \mu_{n - 2}^{\prime} \left( s_{n - 2} - U_{\textrm{ms}} \right) \ .
\end{align*}
The KKT conditions are
\begin{gather*}
  \begin{aligned}
    \frac{\partial L}{\partial s_{i}} = & \ \frac{U_{\textrm{ms}} \cdot \kappa + s_{2} + \dots + s_{i - 1}}{(U_{\textrm{ms}} \cdot \kappa + s_{2} + \dots + s_{i - 1} + s_{i})^{2}} - \sum_{i^{\prime} = i + 1}^{n - 2}{ \frac{s_{i^{\prime}}}{(U_{\textrm{ms}} \cdot \kappa + s_{2} + \dots + s_{i^{\prime}})^2} } \\
                                        & \ - \frac{U_{\textrm{ms}}}{U_{\textrm{ms}} \cdot \kappa + s_{2} + \dots + s_{n - 2} + U_{\textrm{ms}}} - \mu_{i} + \mu_{i}^{\prime}                                                                                                                                       \\
    =                                   & \ 0 \ ,
  \end{aligned} \\
  \left( L_{\textrm{ms}} - s_{i} \right)                  \leq 0 \ , \quad \left( s_{i} - U_{\textrm{ms}} \right)                  \leq 0 \ ,\quad \mu_{i} \leq 0, \quad \mu_{i}^{\prime} \leq 0 \\
  \mu_{i} \left( L_{\textrm{ms}} - s_{i} \right)          = 0 \ , \quad \mu_{i}^{\prime} \left( s_{i} - U_{\textrm{ms}} \right) = 0 \ .
\end{gather*}
We have the KKT point as follows
\[
  (s_{2}, \dots, s_{n - 2}) = (U_{\textrm{ms}}, \dots, U_{\textrm{ms}}) \ .
\]
To verify the KKT point, we first observe that
\begin{align*}
  \sum_{i^{\prime} = i + 1}^{n - 2}{ \frac{1}{(U_{\textrm{ms}} \cdot \kappa + (i^{\prime} - 1)U_{\textrm{ms}})^2} } & + \frac{U_{\textrm{ms}}}{U_{\textrm{ms}} \cdot \kappa + s_{2} + \dots + s_{n - 2} + U_{\textrm{ms}}}                                                                                                        \\
                                                                                                                    & = \sum_{i^{\prime} = i + 1}^{n - 1}{ \frac{1}{(U_{\textrm{ms}} \cdot \kappa + (i^{\prime} - 1)U_{\textrm{ms}})^2} }                                                                                         \\
                                                                                                                    & < \sum_{i^{\prime} = i + 1}^{n - 1}{ \left( \frac{1}{U_{\textrm{ms}} \cdot \kappa + (i^{\prime} - 2) U_{\textrm{ms}}} - \frac{1}{U_{\textrm{ms}} \cdot \kappa + (i^{\prime} - 1) U_{\textrm{ms}}} \right) } \\
                                                                                                                    & = \frac{1}{U_{\textrm{ms}} \cdot \kappa + (i - 1) U_{\textrm{ms}}} - \frac{1}{U_{\textrm{ms}} \cdot \kappa + (n - 2) \cdot U_{\textrm{ms}}} \ ,
\end{align*}
Note that the following holds under Assumption~\ref{assumption:density_ratio}:
\[
  \frac{U_{\textrm{ms}} \cdot \kappa + (i - 2) U_{\textrm{ms}}}{(U_{\textrm{ms}} \cdot \kappa + (i - 1) U_{\textrm{ms}})^{2}} -
  \frac{U_{\textrm{ms}}}{(U_{\textrm{ms}} \cdot \kappa + (i - 1) U_{\textrm{ms}})} +
  \frac{U_{\textrm{ms}}}{(U_{\textrm{ms}} \cdot \kappa + (n - 2) U_{\textrm{ms}})} > 0 \ ,
\]
implying that
\[
  \mu_{i} = 0 \quad \textrm{and} \quad \mu_{i}^{\prime} < 0 \ ,
\]
which verifies the KKT conditions.
Therefore, we have
\[
  \sup_{x_{1: n}} \bar{h}(x_{1}; x_{1: n})
  = \sup_{s_{2}, \dots, s_{n - 1}} \tilde{h}(x_{1}; s_{1}, s_{2}, \dots, s_{n - 1})
  = \tilde{h}(x_{1}; U_{\textrm{ms}} \cdot \kappa, \underbrace{U_{\textrm{ms}}, \dots, U_{\textrm{ms}}}_{n - 2}) \ .
\]
Similarly, we have
\[
  \inf_{x_{1: n}} \bar{h}(x_{n}; x_{1: n})
  = \inf_{s_{1}, \dots, s_{n - 1}} \tilde{h}(x_{n}; s_{1}, s_{2}, \dots, s_{n - 1})
  = \tilde{h}(x_{n}; \underbrace{L_{\textrm{ms}}, \dots, L_{\textrm{ms}}}_{n - 2}, U_{\textrm{ms}}) \ .
\]
By Lemma~\ref{lemma:expected_depth}, we have
\[
  \begin{aligned}
    \sup_{x_{1: n}} \bar{h}(x_{1}; x_{1: n}) & = \tilde{h}(x_{1}; U_{\textrm{ms}} \cdot \kappa, \underbrace{U_{\textrm{ms}}, \dots, U_{\textrm{ms}}}_{n - 2})                                                                                                                                                            \\
                                             & = \frac{U_{\textrm{ms}} \cdot \kappa}{U_{\textrm{ms}} \cdot \kappa} + \frac{U_{\textrm{ms}}}{U_{\textrm{ms}} \cdot \kappa + U_{\textrm{ms}}} + \dots + \frac{U_{\textrm{ms}}}{U_{\textrm{ms}} \cdot \kappa + U_{\textrm{ms}} + \dots + U_{\textrm{ms}} + U_{\textrm{ms}}} \\
                                             & \leq \frac{U_{\textrm{ms}}}{U_{\textrm{ms}}} + \frac{L_\textrm{ms}}{L_{\textrm{ms}} + U_{\textrm{ms}}} + \dots + \frac{L_{\textrm{ms}}}{L_{\textrm{ms}} + \dots + L_{\textrm{ms}} + U_{\textrm{ms}}}                                                                      \\
                                             & = \tilde{h}(x_{n}; \underbrace{L_{\textrm{ms}}, \dots, L_{\textrm{ms}}}_{n - 2}, U_{\textrm{ms}})                                                                                                                                                                         \\
                                             & \leq \inf_{x_{1: n}} \bar{h}(x_{n}; x_{1: n}) \ ,
  \end{aligned}
\]
which completes the proof.
$\hfill\square$

\subsection{Proof of Theorem~\ref{theorem:marginal_single_anomaly_necessity}}
\label{appendix:proof_marginal_single_anomaly_necessity}

We prove by giving the assignment of \( (x_{1}, \dots, x_{n}) \) such that when \( U_{\textrm{ms}} \cdot \kappa > x_{2} - x_{1} > U_{\textrm{ms}} \)
\[
  \exists j > 1, \quad \bar{h}(x_{1}; x_{1: n}) \leq \bar{h}(x_{j}; x_{1: n}) \ .
\]
We assign \( x_{i} \) from the corresponding \( s_{i} \) as follows
\[
  (s_{1}, \dots, s_{n - 1}) = (U_{\textrm{ms}} + \epsilon, U_{\textrm{ms}} / 2, U_{\textrm{ms}}, \dots, U_{\textrm{ms}}) \ ,
\]
where \( \epsilon > 0 \) is relative small than \( U_{\textrm{ms}} \).
By mathematical induction on $n$, we can show that
\begin{align*}
  \tilde{h}(x_{1}; U_{\textrm{ms}}, U_{\textrm{ms}} / 2, U_{\textrm{ms}}, \dots, U_{\textrm{ms}}) & = \tilde{h}(x_{1}; 1, 1 / 2, 1, \dots, 1)                                                         \\
                                                                                                  & > \tilde{h}(x_{n}; 1, 1 / 2, 1, \dots, 1)                                                         \\
                                                                                                  & = \tilde{h}(x_{n}; U_{\textrm{ms}}, U_{\textrm{ms}} / 2, U_{\textrm{ms}}, \dots, U_{\textrm{ms}})
\end{align*}
By the continuity of \( \tilde{h} \) and \( \bar{h} \), there exists an \( \epsilon > 0 \) such that
\[
  \tilde{h}(x_{1}; U_{\textrm{ms}} + \epsilon, U_{\textrm{ms}} / 2, U_{\textrm{ms}}, \dots, U_{\textrm{ms}})
  > \tilde{h}(x_{n}; U_{\textrm{ms}} + \epsilon, U_{\textrm{ms}} / 2, U_{\textrm{ms}}, \dots, U_{\textrm{ms}})
\]
The proof is complete. \(\hfill\square\)

\subsection{Proof of Theorem~\ref{theorem:marginal_single_anomaly_nearest_neighbor}}
\label{appendix:proof_marginal_single_anomaly_nearest_neighbor}

Recall that the output of \( k \)-Nearest Neighbor is
\[
  h_{knn}(x; D) \triangleq \frac{1}{k} \sum_{x^{\prime} \in \mathcal{N}_{k}(x)} \| x - x^{\prime} \|_{1} \ .
\]
Similar to the proof of Theorem~\ref{theorem:marginal_single_anomaly}, we will prove by showing that for all \( j > 1 \), the following holds
\[
  h_{knn}(x_{1}; x_{1: n}) >
  \inf_{x_{1: n}} h_{knn}(x_{1}; x_{1: n}) \geq
  \sup_{j^{\prime}, x_{1: n}} h_{knn}(x_{j^{\prime}}; x_{1: n}) \geq
  h_{knn}(x_{j}; x_{1: n}),
\]
of which the direction of inequalities has reversed direction compared to Theorem~\ref{theorem:marginal_single_anomaly}.
This arises due to the negative correlation between the depth function and the distance to the nearest neighbor.
Similarly, we define
\[
  \tilde{h}_{knn}(x_{1}; s_{1}, \dots, s_{n - 1}) \triangleq h_{knn}(x_{1}; x_{1: n}) \ ,
\]
where \( s_{i} \triangleq x_{i + 1} - x_{1} \) are the difference between two neighbored points.
Then, we have
It is not difficult to show that
\begin{align*}
  \inf_{x_{1: n}} h_{knn}(x_{1}; x_{1: n}) & = \inf_{s_{1: n - 1}} \tilde{h}_{knn}(x_{1}; s_{1: n - 1})                                                                     \\
                                           & = \tilde{h}_{knn}(s_{1}^{\circ}, L_{\textrm{ms}}, \dots, L_{\textrm{ms}})                                                      \\
                                           & = \frac{1}{k} \left[ s_{1}^{\circ} + s_{1}^{\circ} + L_{\textrm{ms}} + \dots + s_{1}^{\circ} + (k - 1) L_{\textrm{ms}} \right] \\
                                           & = \frac{1}{k} \left[ \frac{k(k + 1) s_{1}^{\circ}}{2} + \frac{k(k - 1)}{2} L_{\textrm{ms}} \right]                             \\
                                           & = \frac{k + 1}{2} s_{1}^{\circ} + \frac{k - 1}{2} L_{\textrm{ms}}                                                              \\
                                           & = \frac{k + 1}{2} U_{\textrm{ms}} \ ,
\end{align*}
where \( s_{1}^{\circ} = U_{\textrm{ms}} + (k - 1)(U_{\textrm{ms}} - L_{\textrm{ms}}) / 2 \).
Similarly, we have
\begin{align*}
  \sup_{x_{1: n}} h_{knn}(x_{n}; x_{1: n}) & = \sup_{s_{1: n - 1}} \tilde{h}_{knn}(x_{n}; s_{1: n - 1})                     \\
                                           & = \tilde{h}_{knn}(x_{n}, U_{\textrm{ms}}, \dots, U_{\textrm{ms}})              \\
                                           & = \frac{1}{k} ( U_{\textrm{ms}} + 2U_{\textrm{ms}} + \dots + k U_\textrm{ms} ) \\
                                           & = \frac{k + 1}{2} U_{\textrm{ms}}                                              \\
                                           & = \inf_{x_{1: n}} h_{knn}(x_{1}; x_{1: n}) \ .
\end{align*}
\(\hfill\square\)

\section{Proofs of Theorems~\ref{theorem:central_single_anomaly}-\ref{theorem:central_single_anomaly_nearest_neighbor}}

We will prove Theorems~\ref{theorem:central_single_anomaly}-\ref{theorem:central_single_anomaly_nearest_neighbor} in this section.

\subsection{Proof of Theorem~\ref{theorem:central_single_anomaly}}
\label{appendix:proof_central_single_anomaly}

Before that, we first define \( U_{\textrm{cs}} \) and \( L_{\textrm{cs}} \) as follows:
\[
  U_{\textrm{cs}} = \max_{j \neq n_{0} / 2} \lvert x_{j + 1} - x_{j} \rvert \quad L_{\textrm{cs}} = \min_{j \neq n_{0} / 2} \lvert x_{j + 1} - x_{j} \rvert \ .
\]
Similar to the proof of Theorem~\ref{theorem:marginal_single_anomaly}, we define
\[
  \tilde{h}(x_{i}; s_{1}, \dots, s_{n - 1}) = \bar{h}(x_{i}; x_{1: n}) \ ,
\]
where \( x_{i} \in D \) is an arbitrary element in \( D \).
For convenience, we define
\[
  \theta = \min \{ x_{m + 1} - x_{m}, x_{m + 2} - x_{m + 1} \} \ .
\]
We will prove the sufficiency and necessity of \( \theta = \Theta(\sqrt{n_{0}}) \) separately.

\textbf{Sufficiency:} There exists some constant \( c_{1} \) such that when \( \theta > c_{1} \cdot \sqrt{n_{0}} \), we have
\[
  \forall j, ~ \bar{h}(x_{n_{0} / 2 + 1}; x_{1: n}) < \min_{j} \bar{h}(x_{j}; x_{1: n}) \ .
\]
By the definition of \( \tilde{h} \)
\begin{align*}
  \bar{h}(x_{n_{0} / 2 + 1}; x_{1: n}) & = \tilde{h}(x_{n_{0} / 2 + 1}; s_{1}, \dots, s_{n - 1})                                                                                                                                                                    \\
                                       & < 2 \sup_{s_{1: n_{0} / 2}} \tilde{h}(s_{n_{0} / 2}; s_{1}, \dots, s_{n_{0} / 2})                                                                                                                                          \\
                                       & \leq 2 \tilde{h}( s_{n_{0} / 2}; U_{\textrm{cs}}, \dots, U_{\textrm{cs}}, \theta )                                                                                                                                         \\
                                       & = 2 \left( \frac{\theta}{\theta} + \frac{U_{\textrm{cs}}}{\theta + U_{\textrm{cs}}} + \frac{U_{\textrm{cs}}}{\theta + 2U_{\textrm{cs}}} + \dots + \frac{U_{\textrm{cs}}}{\theta + (n_{0} / 2 - 1) U_{\textrm{cs}}} \right) \\
                                       & \leq 2 + 2 \ln \left[ 1 + \frac{n_{0} / 2 - 1}{\theta / U_{\textrm{cs}}} \right] \ .
\end{align*}
Similarly, we have
\begin{align*}
  \min_{j} \bar{h}(x_{j}; x_{1: n}) & \geq \inf_{j, s_{1: n}} \bar{h}(x_{j}; x_{1: n})                                                                                                                                                                                                       \\
                                    & = \inf_{s_{1: n - 1}} \tilde{h}(x_{1}; s_{1: n - 1}) \textrm{ (follows the analysis in Appendix~\ref{appendix:proof_marginal_single_anomaly})}                                                                                                         \\
                                    & \geq \inf_{s_{1: n_{0} / 2 - 1}} \tilde{h}(x_{1}; s_{1: n_{0} / 2 - 1})                                                                                                                                                                                \\
                                    & = \frac{U_{\textrm{cs}}}{U_{\textrm{cs}}} + \frac{L_{\textrm{cs}}}{U_{\textrm{cs}} + L_{\textrm{cs}}} + \frac{L_{\textrm{cs}}}{U_{\textrm{cs}} + 2L_{\textrm{cs}}} + \dots + \frac{L_{\textrm{cs}}}{U_{\textrm{cs}} + (n_{0} / 2 - 1) L_{\textrm{cs}}} \\
                                    & \geq 1 + \ln \left[ 1 + \frac{n_{0} / 2 - 1}{U_{\textrm{cs}} / L_{\textrm{cs}} + 1} \right] \ .
\end{align*}
Thus, we expect that
\begin{align*}
  2 + 2 \ln \left[ 1 + \frac{n_{0} / 2 - 1}{\theta / U_{\textrm{cs}}} \right]
   & = \ln \left[ e^{2} \left( 1 + \frac{n_{0} / 2 - 1}{\theta / U_{\textrm{cs}}} \right)^{2} \right]         \\
   & \leq \ln \left[ e \left( 1 + \frac{n_{0} / 2 - 1}{U_{\textrm{cs}} / L_{\textrm{cs}} + 1} \right) \right]
  = 1 + \ln \left[ 1 + \frac{n_{0} / 2 - 1}{U_{\textrm{cs}} / L_{\textrm{cs}} + 1} \right] \ ,
\end{align*}
which holds if \( \theta \geq \Omega(\sqrt{n_{0}}) \).

\textbf{Necessity:} If \( \theta < o(\sqrt{n_{0}}) \), there exists some assignment of \( x_{1}, \dots, x_{n} \) such that
\[
  \exists j, ~ \bar{h}(x_{n_{0} / 2 + 1}; x_{1: n}) \geq \bar{h}(x_{j}; x_{1: n}) \ .
\]
We prove this by giving an example of such an assignment as follows:
\[
  x_{1} = 1, \quad x_{2} = 2, \quad \dots, \quad x_{n_{0} / 2} = n_{0} / 2, \quad x_{n_{0} / 2 + 1} = n_{0} / 2 + \theta,
\]
\[
  x_{n_{0} / 2 + 2} = n_{0} / 2 + 2\theta, \quad x_{n_{0} / 2 + 3} = n_{0} / 2 + 2\theta + 1, \quad \dots, \quad x_{n} = n_{0} / 2 + 2 \theta + n_{0} / 2 \ ,
\]
or equivalently,
\[
  (s_{1}, \dots, s_{n - 1}) = \left( 1, \dots, 1, \theta, \theta, 1, \dots, 1 \right) \ .
\]
Then we have
\begin{align*}
  \tilde{h}(x_{n_{0} / 2 + 1}) & = 2 \left( \frac{\theta}{\theta} + \frac{1}{\theta + 1} + \frac{1}{\theta + 2} + \dots + \frac{1}{\theta + (n_{0} / 2 - 1)} \right) \\
                               & \geq 2 + 2 \ln \left( 1 + \frac{n_{0} / 2 - 1}{\theta + 1} \right)                                                                  \\
                               & \geq \ln \left( \Omega(n_{0}^{2} / \theta^{2}) \right)
\end{align*}
and
\begin{align*}
  \tilde{h}(x_{1}) = & \  1 + \frac{1}{2} + \frac{1}{3} + \dots + \frac{1}{n_{0} / 2 - 1} + \frac{\theta}{n_{0} / 2 - 1 + \theta} +                                  \\
                     & \  \frac{\theta}{2\theta + n_{0} / 2 - 1} + \frac{1}{n_{0} / 2 - 1 + 2\theta + 1} + \dots + \frac{1}{n_{0} / 2 - 1 + 2\theta + n_{0} / 2 - 1} \\
  \leq               & \  1 + \frac{1}{2} + \frac{1}{3} + \dots + \frac{1}{n_{0} / 2 - 1} + 2 + \ln(2)                                                               \\
  \leq               & \  3 + \ln(2) + \ln \left( n_{0} / 2 - 1 \right)                                                                                              \\
  =                  & \  \ln(O(n_{0}))
\end{align*}
Therefore, if \( \theta = o(\sqrt{n_{0}}) \), we have
\[
  \tilde{h}(x_{n_{0} / 2 + 1}) = \omega(\tilde{h}(x_{1})) \ ,
\]
implying that when \( n_{0} \) is large enough
\[
  \bar{h}(x_{n_{0} / 2 + 1}; x_{1: n}) > \bar{h}(x_{1}; x_{1: n}) \ .
\]
$\hfill\square$

\subsection{Proof of Theorem~\ref{theorem:central_single_anomaly_nearest_neighbor}}
\label{appendix:proof_central_single_anomaly_nearest_neighbor}

For convenience, we let \( k \) be an even number, otherwise, choose \( k + 1 \) instead of \( k \) acts similarly.
Following the proof of Theorem~\ref{theorem:marginal_single_anomaly_nearest_neighbor}, we have
\begin{align*}
  \min_{s_{1: n - 1}}\tilde{h}_{knn}(x_{n_{0} / 2 + 1}; s_{1}, \dots, s_{n - 1}) & = \tilde{h}_{knn}(x_{n_{0} / 2 + 1}; \underbrace{L_{\textrm{cs}}, \dots, L_{\textrm{cs}}}_{n_{0} / 2 - 1}, \theta, \theta, \underbrace{L_{\textrm{cs}}, \dots, L_{\textrm{cs}}}_{n_{0} / 2 - 1}) \\
                                                                                 & = \frac{2}{k} \left( \theta + \theta + L_{\textrm{cs}} + \dots \theta + (k / 2 - 1) L_{\textrm{cs}} \right)                                                                                      \\
                                                                                 & = \frac{2}{k} \left( k\theta / 2 + \frac{k / 2 (k / 2 - 1)}{2} L_{\textrm{cs}} \right)                                                                                                           \\
                                                                                 & = \theta + \frac{k / 2 - 1}{2} L_{\textrm{cs}} \ .
\end{align*}
Similarly, we have
\begin{align*}
  \max_{s_{1: n - 1}}\tilde{h}_{knn}(x_{1}; s_{1}, \dots, s_{n - 1}) & = \tilde{h}_{knn}(x_{1}; \underbrace{U_{\textrm{cs}}, \dots, U_{\textrm{cs}},}_{n_{0} / 2 - 1} \theta, \theta, \underbrace{U_{\textrm{cs}}, \dots, U_{\textrm{cs}},}_{n_{0} / 2 - 1}) \\
                                                                     & = \frac{1}{k} \left( U_{\textrm{cs}} + 2 U_{\textrm{cs}}\dots + k U_{\textrm{cs}} \right)                                                                                             \\
                                                                     & = \frac{k + 1}{2} U_{\textrm{cs}} \ .
\end{align*}
Then, when \( \theta > \frac{k + 1}{2} U_{\textrm{cs}} - \frac{k / 2 - 1}{2} L_{\textrm{cs}} = \Omega(k) \), we have \( \tilde{h}_{knn}(x_{n_{0} / 2 + 1}) > \max_{j} \tilde{h}_{knn}(x_{j}) \).
The necessity can be proven similarly as in Appendix~\ref{appendix:proof_central_single_anomaly}, by giving an example of such an assignment as follows:
\[
  (s_{1}, \dots, s_{n - 1}) = ( \underbrace{1, \dots, 1}_{n_{0} / 2 - 1}, \theta, \theta, \underbrace{1, \dots, 1}_{n_{0} / 2 - 1} ) \ .
\]
$\hfill\square$

\section{Proofs of Theorems~\ref{theorem:marginal_multiple_anomalies}-\ref{theorem:marginal_multiple_anomalies_nearest_neighbor}}

Here, We will prove Theorems~\ref{theorem:marginal_multiple_anomalies}-\ref{theorem:marginal_multiple_anomalies_nearest_neighbor}.
We begin with the following definitions.
\[
  U_{\textrm{mg}} = \max_{1 \leq j \leq n_{1} - 1} \lvert x_{j + 1} - x_{j} \rvert, \quad L_{\textrm{mg}} = \min_{n_{1} + 1 \leq j \leq n_{1} + n_{0} - 1} \lvert x_{j + 1} - x_{j} \rvert, \quad \textrm{and } \theta = x_{n_{1} + 1} - x_{n_{1}} \ .
\]
As we assume that \( n_{1} \) is odd, we have that \( n_{1} = 2q + 1 \) for some \( q \in N^{+} \).

\subsection{Proof of Theorem~\ref{theorem:marginal_multiple_anomalies}}
\label{appendix:proof_marginal_multiple_anomalies}

\textbf{Sufficiency:} There exists some constant \( c_{1} \) such that when \( \theta \geq c_{1} \cdot n_{1}^{2} \), we have
\[
  \max_{1 \leq j \leq n_{1}} \bar{h}(x_{j}; x_{1: n}) \leq \min_{n_{1} + 1 \leq j \leq n_{1} + n_{0}} \bar{h}(x_{j}; x_{1: n}) \ .
\]
Following the analysis in Appendix~\ref{appendix:proof_marginal_single_anomaly}, we have
\[
  \max_{1 \leq j \leq n_{1}}
  \bar{h}(x_{j}; x_{1: n})
  = \bar{h}(x_{q + 1}; x_{1: n})
  \quad \textrm{and} \quad
  \min_{2q + 2 \leq j \leq 2q + 1 + n_{0}} \bar{h}(x_{j}; x_{1: n})
  = \bar{h}(x_{n}; x_{1: n}) \ .
\]
Then, it suffices to show that
\[
  \sup_{x_{1: n}}
  \bar{h}(x_{q + 1}; x_{1: n})
  \leq
  \inf_{x_{1: n}}
  \bar{h}(x_{n}; x_{1: n}) \ .
\]
Similar to the analysis in Appendix~\ref{appendix:proof_marginal_single_anomaly}, we have
\begin{align*}
  \sup_{x_{1: n}}
  \bar{h}(x_{2q + 1}; x_{1: n})
   & =
  \sup_{s_{1: q}} \tilde{h}(x_{2q + 1}; s_{1: q}) +
  \sup_{s_{q + 1: 2q + n_{0}}} \tilde{h}(x_{2q + 1}; s_{q + 1: 2q + n_{0}})                                                                                                                                                                                    \\
   & =
  \tilde{h}(x_{q + 1}; \underbrace{U_{\textrm{mg}}, \dots, U_{\textrm{mg}}}_{q - 1}, L_{\textrm{mg}}, L_{\textrm{mg}}, \underbrace{U_{\textrm{mg}}, \dots, U_{\textrm{mg}}}_{q - 1}, \theta, \underbrace{U_{\textrm{mg}}, \dots, U_{\textrm{mg}}}_{n_{0} - 2}) \\
   & =
  2 \left(
  \frac{L_{\textrm{mg}}}{L_{\textrm{mg}}} +
  \frac{U_{\textrm{mg}}}{L_{\textrm{mg}} + U_{\textrm{mg}}} +
  \dots +
  \frac{U_{\textrm{mg}}}{L_{\textrm{mg}} + (q - 1) U_{\textrm{mg}}}
  \right) +                                                                                                                                                                                                                                                    \\
   & \quad\ \frac{\theta}{2 L_{\textrm{mg}} + 2(q - 1) U_{\textrm{mg}} + \theta} +                                                                                                                                                                             \\
   & \quad\ \frac{U_{\textrm{mg}}}{2 L_{\textrm{mg}} + 2(q - 1) U_{\textrm{mg}} + \theta + U_{\textrm{mg}}} +                                                                                                                                                  \\
   & \quad\ \dots +                                                                                                                                                                                                                                            \\
   & \quad\ \frac{U_{\textrm{mg}}}{2 L_{\textrm{mg}} + 2(q - 1) U_{\textrm{mg}} + \theta + (n_{0} - 2) U_{\textrm{mg}}}                                                                                                                                        \\
   & \leq \ln O \left(  q^{2} n_{0} / (q + \theta) \right) \ .
\end{align*}
and
\begin{align*}
  \inf_{x_{1: n}}
  \bar{h}(x_{n}; x_{1: n})
   & =
  \inf_{s_{1: n - 1}} \tilde{h}(x_{n}; s_{1: n - 1})                                                                                                                     \\
   & =
  \tilde{h}(x_{n}; \underbrace{L_{\textrm{mg}}, \dots, L_{\textrm{mg}}}_{2q}, \theta, \underbrace{L_{\textrm{mg}}, \dots, L_{\textrm{mg}}}_{n_{0} - 2}, U_{\textrm{mg}}) \\
   & =
  \frac{U_{\textrm{mg}}}{U_{\textrm{mg}}} +
  \frac{L_{\textrm{mg}}}{U_{\textrm{mg}} + L_{\textrm{mg}}} +
  \dots +
  \frac{L_{\textrm{mg}}}{U_{\textrm{mg}} + (n_{0} - 2) U_{\textrm{mg}}} + * \quad                                                                                        \\
   & \geq \ln \left( \Omega \left( n_{0} \right) \right) \ ,
\end{align*}
where \(*\) is some positive quantity. Therefore, \( \theta = \Omega(q^{2}) = \Omega(n_{1}^{2}) \) suffices to ensure that
\[
  \sup_{x_{1: n}}
  \bar{h}(x_{q + 1}; x_{1: n})
  \leq
  \inf_{x_{1: n}}
  \bar{h}(x_{n}; x_{1: n}) \ ,
\]
implying that
\[
  \max_{1 \leq j \leq n_{1}}
  \bar{h}(x_{j}; x_{1: n})
  \leq
  \min_{n_{1} + 1 \leq j \leq n_{1} + n_{0}}
  \bar{h}(x_{j}; x_{1: n}) \ .
\]

\textbf{Neccessity:} We will show that if \( \theta = o(n_{1}^{2}) \), then there exists some assignment of \( x_{1: n} \) such that
\[
  \max_{1 \leq j \leq n_{1}}
  \bar{h}(x_{j}; x_{1: n})
  >
  \min_{n_{1} + 1 \leq j \leq n_{1} + n_{0}}
  \bar{h}(x_{j}; x_{1: n}) \ .
\]
We similarly give the assignment of \( x_{1: n} \) as follows:
\[
  (s_{1}, \dots, s_{n_{1} - 1}) = (\underbrace{1, \dots, 1}_{2q}, \theta, \underbrace{1, \dots, 1}_{n_{0} - 1}) \ .
\]
Then, we have
\begin{align*}
  \tilde{h}(x_{q + 1}; s_{1: n - 1}) & = 2\left( 1 + \frac{1}{2} + \dots \frac{1}{q} \right) + \frac{\theta}{2q + \theta} + \frac{1}{2q + \theta + 1} + \dots + \frac{1}{2q + \theta + n_{0} - 2} \\
                                     & \geq \ln \left( \Omega \left( q^{2} n_{0} / (q + \theta) \right) \right)                                                                                   \\
\end{align*}
and
\begin{align*}
  \tilde{h}(x_{n}; s_{1: n - 1}) = & \  1 + \frac{1}{2} + \frac{1}{3} + \dots + \frac{1}{n_{0} - 2} + \frac{\theta}{n_{0} - 2 + \theta + 1}             \\
                                   & + \frac{1}{n_{0} - 2 + \theta + 2} + \dots + \frac{1}{n_{0} - 2 + \theta + 2q} \leq \ln O \left( n_{0} \right) \ .
\end{align*}
Thus, if \( \theta = o(n_{1}^{2}) \), then
\[
  \max_{1 \leq j \leq n_{1}}
  \bar{h}(x_{j}; x_{1: n})
  >
  \min_{n_{1} + 1 \leq j \leq n_{1} + n_{0}}
  \bar{h}(x_{j}; x_{1: n}) \ .
\]
The proof is complete. \( \hfill\square \)

\subsection{Proof of Theorem~\ref{theorem:marginal_multiple_anomalies_nearest_neighbor}}
\label{appendix:proof_marginal_multiple_anomalies_nearest_neighbor}

\textbf{Sufficiency:} There exists some constant \( c_{1} \) such that when \( \theta \geq c_{1} k \), we have
\[
  \min_{1 \leq j \leq n_{1}} h_{knn}(x_{j}; x_{1: n}) \geq \max_{n_{1} + 1 \leq j \leq n_{1} + n_{0}} h_{knn}(x_{j}; x_{1: n}) \ .
\]
Observing that
\begin{align*}
  \min_{1 \leq j \leq n_{1}} h_{knn}(x_{j}; x_{1: n}) & = \inf_{1 \leq j \leq n_{1}, s_{1: n - 1}} \tilde{h}(x_{j}; s_{1: n - 1})                                                                                   \\
                                                      & = \inf_{1 \leq j \leq n_{1}, s_{1: n_{1} - 1}} \tilde{h}(x_{j}; s_{1: n_{1} - 1}, \theta, \underbrace{L_{\textrm{mg}}, \dots, L_{\textrm{mg}}}_{n_{0} - 1}) \\
                                                      & \geq \frac{1}{k} \left[ (k - n_{1} - 2) \theta + (k - n_{1} - 2)(k - n_{1} - 3) L_{\textrm{mg}} / 2  \right]                                                \\
                                                      & = \Omega(\theta + k)
\end{align*}
and
\begin{align*}
  \max_{n_{1} + 1 \leq j \leq n_{1} + n_{0}} h_{knn}(x_{j}; x_{1: n}) & = \sup_{n_{1} + 1 \leq j \leq n_{1} + n_{0}, s_{1: n - 1}} \tilde{h}(x_{j}; s_{1: n - 1})                                                                                   \\
                                                                      & = \sup_{n_{1} + 1 \leq j \leq n_{1} + n_{0}, s_{1: n_{1} - 1}} \tilde{h}(x_{n}; s_{1: n_{1} - 1}, \theta, \underbrace{U_{\textrm{mg}}, \dots, U_{\textrm{mg}}}_{n_{0} - 1}) \\
                                                                      & \leq \frac{1}{k} \left( U_{\textrm{mg}} + 2 U_{\textrm{mg}} + \dots + k U_{\textrm{mg}} \right)                                                                             \\
                                                                      & = O(k) \ .
\end{align*}
Therefore, \( \theta = \Omega(k) \) suffices to ensure that
\[
  \min_{1 \leq j \leq n_{1}} h_{knn}(x_{j}; x_{1: n}) \geq \max_{n_{1} + 1 \leq j \leq n_{1} + n_{0}} h_{knn}(x_{j}; x_{1: n}) \ .
\]

\textbf{Neccessity:} We will show that if \( \theta = o(k) \), then there exists some assignment of \( x_{1: n} \) such that
\[
  \min_{1 \leq j \leq n_{1}}
  h_{knn}(x_{j}; x_{1: n})
  <
  \max_{n_{1} + 1 \leq j \leq n_{1} + n_{0}}
  h_{knn}(x_{j}; x_{1: n}) \ .
\]
We similarly give the assignment of \( x_{1: n} \) corresponding to \( s_{1: n - 1} \) as follows:
\[
  (s_{1}, \dots, s_{n - 1}) = (\underbrace{L_{\textrm{mg}}, \dots, L_{\textrm{mg}}}_{n_{1} - 1}, \theta, \underbrace{L_{\textrm{mg}}, \dots, L_{\textrm{mg}}}_{k - n_{1} - 2}, \underbrace{U_{\textrm{mg}}, \dots, U_{\textrm{mg}}}_{n - k + 1}) \ .
\]
Then, we have
\begin{align*}
  \min_{1 \leq j \leq n_{1}} h_{knn}(x_{j}; x_{1: n}) & \leq \frac{1}{k} \Big( L_{\textrm{mg}} + 2 L_{\textrm{mg}} + \dots + (n_{1} - 1) L_{\textrm{mg}} + \theta + (\theta + L_{\textrm{mg}}) + \dots + (\theta + (k - n_{1}) L_{\textrm{mg}}) \Big) \\
                                                      & \leq \left[ \frac{n_{1}(n_{1} - 1)}{2} + \frac{(k - n_{1})(k - n_{1} - 1)}{2} \right] \frac{L_{\textrm{mg}}}{k} + \theta                                                                      \\
\end{align*}
and
\begin{align*}
  \max_{n_{1} + 1 \leq j \leq n_{1} + n_{0}} h_{knn}(x_{j}; x_{1: n}) & \geq \frac{1}{k}(1 + 2 + \dots + k) U_{\textrm{mg}} \ .
\end{align*}
Thus, we have
\[
  \min_{1 \leq j \leq n_{1}} h_{knn}(x_{j}; x_{1: n}) - \max_{n_{1} + 1 \leq j \leq n_{1} + n_{0}} h_{knn}(x_{j}; x_{1: n}) \leq -\Omega(k) + \theta \ .
\]
Once \( \theta = o(k) \), we have
\[
  \min_{1 \leq j \leq n_{1}} h_{knn}(x_{j}; x_{1: n}) - \max_{n_{1} + 1 \leq j \leq n_{1} + n_{0}} h_{knn}(x_{j}; x_{1: n}) \leq -\Omega(k) < 0 \ .
\]
The proof is complete. \( \hfill\square \)

\end{document}